\documentclass[10pt,journal,compsoc]{IEEEtran}
\usepackage{epsfig}
\usepackage{graphicx}
\usepackage{amsmath}
\usepackage{amssymb}
\usepackage{tabularx}
\usepackage{subcaption}
\usepackage{xcolor}
\usepackage{threeparttable}
\usepackage{algorithm}
\usepackage{algorithmic}
\usepackage{enumitem}
\usepackage{multirow}
\usepackage{lipsum}
\usepackage[hidelinks,pagebackref=false,breaklinks=true,colorlinks,bookmarks=false]{hyperref}
\ifCLASSOPTIONcompsoc
  \usepackage[nocompress]{cite}
\else
  \usepackage{cite}
\fi
\ifCLASSINFOpdf
\else
\fi

\usepackage{comment}

\usepackage{xspace}
\makeatletter
\DeclareRobustCommand\onedot{\futurelet\@let@token\@onedot}
\def\@onedot{\ifx\@let@token.\else.\null\fi\xspace}

\newcolumntype{P}[1]{>{\centering\arraybackslash}p{#1}}

\makeatletter
\renewcommand*{\p@section}{\S\,}
\renewcommand*{\p@subsection}{\S\,}
\makeatother

\newcommand{\erhao}[1]{\fontsize{11pt}{\baselineskip}\selectfont}

\usepackage{url}
\usepackage[utf8]{inputenc}
\usepackage{booktabs}
\usepackage[american]{babel}
\usepackage[T1]{fontenc}
\usepackage{balance}

\begin{document}
\title{The State of Aerial Surveillance: A Survey}
%\title{Deep Complex-valued Iris Network: \\Bridging the Gap Between Deep Learning and Iris Recognition}
%\title{Deep Complex-valued Network \\for Iris Recognition}
%\title{Optimal Feature Representation \\for Iris Recognition}

\author{K.~Nguyen, ~\IEEEmembership{Member,~IEEE,}
        C.~Fookes, ~\IEEEmembership{Senior Member,~IEEE,}
        S.~Sridharan, ~\IEEEmembership{Life Senior Member,~IEEE,}  
        Y.~Tian, ~\IEEEmembership{Fellow,~IEEE,}
        F.~Liu, %~\IEEEmembership{Member,~IEEE,}
        X.~Liu, ~\IEEEmembership{Senior Member,~IEEE,}
        A.~Ross, ~\IEEEmembership{Senior Member,~IEEE}
                % <-this % stops a space
\IEEEcompsocitemizethanks{\IEEEcompsocthanksitem K. Nguyen, C. Fookes and S. Sridharan are with the Research Program in Signal Processing, AI and Vision Technologies (SAIVT), Queensland University of Technology, Australia. \protect 
E-mail: \{k.nguyenthanh,c.fookes,s.sridharan\}@qut.edu.au}% <-this % stops an unwanted space

%\IEEEcompsocitemizethanks{\IEEEcompsocthanksitem Simon Lucey is with Australian Institute for Machine Learning, University of Adelaide, Australia. \protect 
%E-mail: simon.lucey@adelaide.edu.au.}

\IEEEcompsocitemizethanks{\IEEEcompsocthanksitem Yingli Tian is with The City College of New York, United States. \protect 
E-mail: ytian@ccny.cuny.edu}

\IEEEcompsocitemizethanks{\IEEEcompsocthanksitem Xiaoming Liu, Feng Liu and Arun Ross are with Department of Computer Science and Engineering, Michigan State University, United States. \protect 
E-mail: \{liuxm,liufeng6,rossarun\}@cse.msu.edu.}

%\thanks{Manuscript received April 8th, 2019}
}


%\markboth{IEEE Transactions on Pattern Analysis and Machine Intelligence}%
\markboth{IEEE Transactions}%
{Nguyen \MakeLowercase{\textit{et al.}}: Aerial Surveillance: A Survey}

\IEEEtitleabstractindextext{%
\begin{abstract}
The rapid emergence of airborne platforms and imaging sensors are enabling new forms of aerial surveillance due to their unprecedented advantages in scale, mobility, deployment and covert observation capabilities. This paper provides a comprehensive overview of human-centric aerial surveillance tasks from a computer vision and pattern recognition perspective. It aims to provide readers with an in-depth systematic review and technical analysis of the current state of aerial surveillance tasks using drones, UAVs and other airborne platforms. The main object of interest is humans, where single or multiple subjects are to be detected, identified, tracked, re-identified and have their behavior analyzed. More specifically, for each of these four tasks, we first discuss unique challenges in performing these tasks in an aerial setting compared to a ground-based setting. We then review and analyze the aerial datasets publicly available for each task, and delve deep into the approaches in the aerial literature and investigate how they presently address the aerial challenges. We conclude the paper with discussion on the missing gaps and open research questions to inform future research avenues.

%R. Kumar et al., "Aerial video surveillance and exploitation," in Proceedings of the IEEE, vol. 89, no. 10, pp. 1518-1539, Oct. 2001, doi: 10.1109/5.959344.

%Paper on Satellites for Wide Area Surveillance Australia
%https://airpower.airforce.gov.au/APDC/media/PDF-Files/Fellowship%20Papers/FELL04-The-Potential-of-Satellites-for-Wide-Area-Surveillance-of-Australia.pdf

%To be read: https://www.technologyreview.com/2019/06/26/102931/satellites-threaten-privacy/?utm_medium=tr_social&utm_campaign=site_visitor.unpaid.engagement&utm_source=Facebook&fbclid=IwAR1Djz5U-S_2fYGn9yDzFYnE98pkwq6jcuUiwnt3SXgMBqNg8vnETOSzIM0#Echobox=1590508801

\end{abstract}

}

\maketitle

\IEEEdisplaynontitleabstractindextext

\IEEEpeerreviewmaketitle

\IEEEraisesectionheading{\section{Introduction}
\label{sec:introduction}}
\IEEEPARstart{T}{his} paper presents the first ever review of the state-of-the-art research of the rapidly evolving area of aerial surveillance, covering the unique challenges in performing the human centric tasks of detection, tracking, recognition, person re-detection and action recognition on aerial data. Large volumes of human centric aerial surveillance data are being collected in several major initiatives. For example in 2011, the Gorgon Stare project, led by the Pentagon in the United States, rolled out a MQ-9 drone equipped with an advanced Multi-Spectral Targeting System called ARGUS to Iraq and Afghanistan. With as many as 368 individual cameras, Gorgon Stare could capture 1.8 billion pixels per frame, enough imaging power to spot an object six inches wide from an altitude of 25,000 feet \cite{EyesInTheSky}. An entire city of size $10 \times 10$ km$^2$ could be continuously observed at a resolution sufficient to monitor any person or vehicle in the city. This all-seeing system enabled persistent and mass surveillance of a wide area 24/7 without awareness of the citizens in the city. 

In 2019, a Stratollite - a giant stratospheric balloon, orbited continuously for 45 days in the stratosphere at an altitude of 65,000 feet to monitor North America \cite{BalloonSurveillance}. Equipped with hi-tech radars and multi-spectral sensors, these solar-powered balloons can hover over a small area of interest, taking photos with a quality of five centimeters per pixel and beaming the footage down to the ground station. This resolution is many times superior than commercial satellites and is sufficient to detect a mobile phone in a person's hand and track all individual vehicles day and night in an entire city for weeks, even months \cite{BalloonSurveillance}.
% Solar-powered UAVs in Stratollite, flying up to 1 year: Airbus Zephyr & UK/Aus DSTG PHASA-35

In the last four consecutive years, four "Vision meets drones" challenges in major academic conferences (ECCV 2018, ICCV 2019, ECCV 2020, ICCV 2021) were organized targeting aerial object detection, tracking and crowd counting from drone-based footage. The challenges were based on a large-scale aerial dataset, VisDrone, with more than 400 aerial videos formed by 265K frames and 2.6M bounding boxes or points of targets of frequent interests, such as pedestrians, cars, and bicycles \cite{VisDrones}. Captured by the civilian DJI Mavic drones flying at an altitude of a few hundred feet in urban areas, the VisDrone footage shows the potential and feasibility of large data collection for analysis.
%an off-the-shelf and cheap aerial surveillance solution for anyone. 

\begin{figure*}
    \centering
    \includegraphics[width=2\columnwidth]{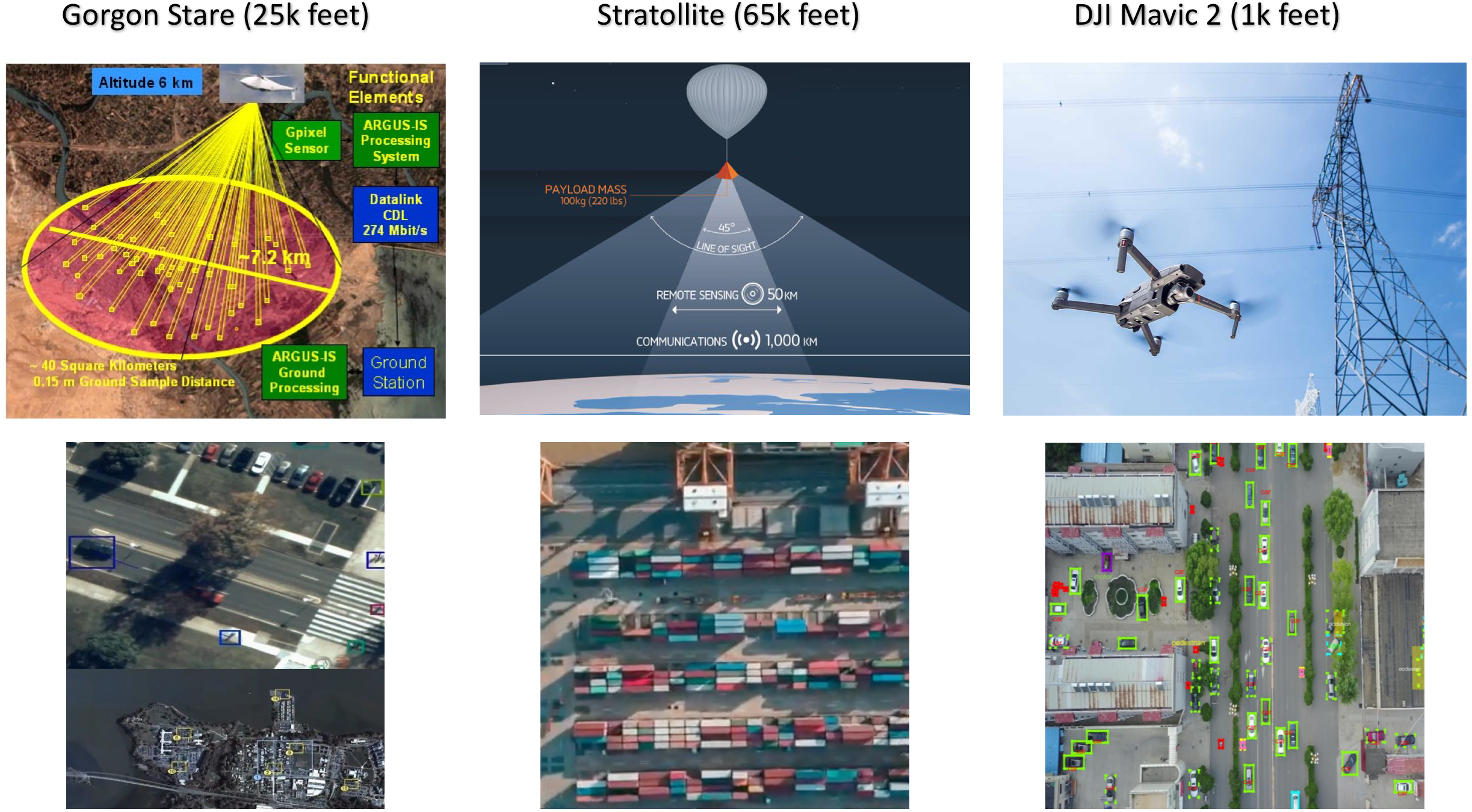}
    \caption{The rise of aerial surveillance. (1) The first column shows the military all-seeing Gorgon Stare \cite{EyesInTheSky} deployed in Iraq, using a MQ-9 drone (costs \$17M USD) flying at an altitude of $25K$ feet, able to capture simultaneously an area of $10 \times 10$ km, with a 1.8 GB sensor at a Ground Sampling Distance (GSD) of 15 cm. (2) The second column is Stratollite, a giant solar-powered balloon flying at an altitude of $65K$ feet \cite{BalloonSurveillance}, which can hover for months, able to detect a mobile phone in a person's hand at a GSD of 5 cm. (3) The third column is the off-the-shelf DJI Mavic 2 (costs \$3K USD) flying at an altitude of $1K$ feet, able to capture simultaneously an area of $1 \times 1$ km, at a GSD of 20 cm \cite{VisDrones}.}
    \label{fig:AerialSurveillance}
\end{figure*}{}

\begin{figure}
    \centering
    \includegraphics[width=0.9\columnwidth]{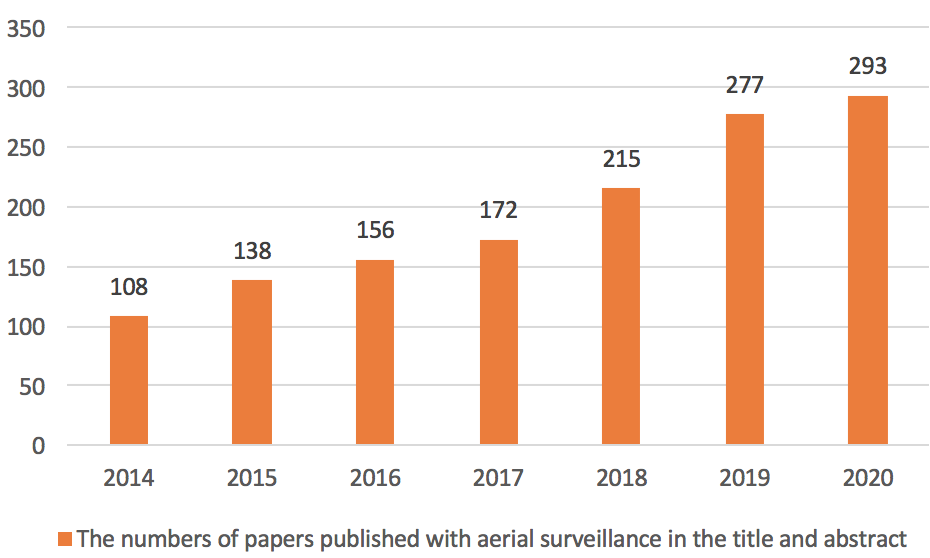}
    \caption{According to Scopus, there are more than 1,300 papers published with one aerial keyword, \emph{i.e.} UAV or Drone or Aerial, and one surveillance keyword, \emph{i.e.} detection or tracking or identification or action or event in the title and abstract in the last six years. }
    \label{fig:AerialSurveillancePapers}
\end{figure}

From the all-seeing eyes of the Gorgon Stare and the months-lasting Stratollites, to the off-the-shelf DJI drones, as illustrated in Fig.~\ref{fig:AerialSurveillance}, the uptake of aerial surveillance has significantly broadened across military, industry and academic sectors.
%an improvised explosive device (IED) tracking in battle front lines to civilian property monitoring, from mountainous border patrol to urban traffic monitoring, from maritime surveillance, drug trafficking detection to search and rescue tasks. 
The academic research related to aerial surveillance has seen a huge boom. According to Scopus, there are more than 78K UAV/drone/aerial papers published \footnote{ \emph{Query string:} TITLE-ABS (uav OR drone OR aerial)}, 38K surveillance papers published \footnote{\emph{Query string:} TITLE-ABS (surveillance) AND TITLE-ABS (detection OR tracking OR identification OR action OR event)} and 1.3K UAV/drone/aerial surveillance papers published \footnote{\emph{Query string:} TITLE-ABS (uav  OR  drone  OR  aerial) AND TITLE-ABS (surveillance) AND TITLE-ABS (detection OR tracking OR identification OR action OR event)} in the last six years as illustrated in Fig.~\ref{fig:AerialSurveillancePapers}.

%\belowcaptionskip

Powered by recent breakthroughs in computer vision and deep learning, the implementation of the basic aerial surveillance tasks of detection, tracking, identification, and action recognition, on aerial data are emerging as an important and timely research area to be investigated. 
%The key advantage of aerial surveillance is that the chance that any person, any vehicle or any area is being unblinkingly watched 24/7. The rapid emerging of airborne platforms and sensors, coupled with recent breakthroughs in computer vision and machine learning, has made aerial surveillance a significant and timely research area to be investigated. Examples of three systems are illustrated in Fig.~\ref{fig:AerialSurveillance}. 
%The increasing deployment from military to civilian, from high-end aircraft platforms to as cheap as \$2000 platforms
%Compared with ground-based surveillance, aerial surveillance provides unprecedented advantages in surveillance scale, mobility and intrusiveness. However, it also exhibits extreme challenges in object viewing angles and object resolutions with diverse backgrounds from moving cameras.
This paper dives deep into aerial surveillance tasks from a computer vision and pattern recognition perspective. Compared with ground-based surveillance, aerial surveillance provides unprecedented scope in surveillance scale, mobility and covert observation. However, it also poses distinct challenges for the computer vision and deep learning community to address including extreme object viewing angles, low-resolution of objects and non-linear object distribution with diverse backgrounds from moving cameras. In surveillance, it is of great interest to detect and track humans, identify and re-identify humans from a camera or across multiple cameras, understand their behaviors and interactions, as well as search for humans with specific characteristics. From a computer vision and machine learning perspective, these tasks are categorized as detection, identification (biometrics, soft-biometrics, and re-ID), action recognition and crowd analysis. While these tasks have been actively studied in generic ground-based images, the challenges with aerial images are daunting but targeted research to address these challenges are rapidly emerging in the aerial domain. This paper will investigate each task from 4 angles: 
\begin{itemize}
    %\item (1) \underline{\textit{ground-based literature:}} review modern trends and approaches in generic ground-based literature, 
    \item (1) \underline{\textit{aerial surveillance challenges:}} discuss unique challenges for performing the task in the aerial domain, and the amount of performance drop when ground-based approaches shift to the aerial domain,
    \item (2) \underline{\textit{aerial surveillance datasets:}} analyze the aerial datasets publicly available for the task,
    \item (3) \underline{\textit{approaches to solve aerial surveillance challenges:}} delve deep in the state-of-the-art approaches in the aerial literature and how they address the challenges unique to aerial data,
    \item (4) \underline{\textit{techniques to improve aerial surveillance tasks:}} further investigate techniques to improve the performance of the aerial task.
\end{itemize}
The rest of the paper is structured as follows. Section~\ref{sec:HighLevelOverview} identifies the advantages and challenges of aerial surveillance, and the statistics of public datasets and academic papers published. Sections~\ref{sec:aerialobjectdetection} -~\ref{sec:aerialactionrecognition} review the state-of-the-art research of the four central aerial surveillance tasks: detection, tracking, identification, and action recognition. Section~\ref{sec:GapsOutlook} summarizes the state of aerial surveillance, open questions and future outlook. Section~\ref{sec:Conclusion} concludes the paper.

    %\item Section~\ref{sec:aerialsceneunderstanding}: 3D scene reconstruction, image stitching, place recognition, quality measurement and denoising (Image stabilization, atmospheric turbulence, motion blur)} ----- We may need to skip this due to the 20-pages limit of PAMI.

\section{Aerial surveillance}
\label{sec:HighLevelOverview}
Aerial surveillance is the task of employing airborne platforms and imaging sensors mounted on them to detect, track, identify, and monitor behavior and activities of a person or a group of persons \cite{EyesInTheSky,VisDrones,UAV-Human}. Aerial surveillance is also known as Wide Area Motion Imagery (WAMI) or mass surveillance since it is capable of monitoring a wide range area, such as an entire city with a large population persistently and unblinkingly 24/7. A wide range of airborne platforms are available with diverse characteristics in flying ranges and altitudes, endurance, speed, manoeuvrability, payload and vulnerabilities as summarized in Table~\ref{tab:airborneplatforms} in Appendix. These airborne platforms are deployed with a wide range of sensors. In this paper, we consider mainly techniques for analyzing data acquired by RGB imaging sensors, which capture data with spatial details to be analyzed by computer vision and deep learning. Imaging sensors are categorized by their spectrum as summarized in Table~\ref{tab:aerialsensors} in Appendix.

To limit the scope of this paper, we only focus on the human-centered security-related surveillance tasks. Aerial surveillance empowers a wide range of human-centric applications, including border patrol \cite{BorderiSurveillance2}, search and rescue \cite{SearchRescureThermal,HERIDAL}, maritime surveillance \cite{MaritimeSurveillance}, protest monitoring, drug trafficking monitoring \cite{DrugTraffickSurveillance}, military IED tracking \cite{EyesInTheSky} and crime fighting \cite{CrimeFighting}. Due to these burgeoning applications of aerial surveillance, corporate aerial surveillance is rapidly growing. Along with this growth, there is a growing privacy threat. Aerial surveillance of individuals as well as mass surveillance can now be conducted at low cost using the drone technology. This poses a serious threat to privacy since there are not well-established privacy protections to prevent widespread and indiscriminate aerial surveillance \cite{SurveillancePrivacy}. In this review we will not be covering the privacy issues of aerial surveillance which is a very important topic. Since aerial surveillance has immense potential to provide safety and security to the public, it is important to ensure that the regulations that are imposed to protect privacy do not hamper the development of the area of aerial surveillance.

\vspace{-3px}
\subsection{Advantages and challenges}
\label{sec:AdvantagesChallenges}
Aerial surveillance can be employed independently or complementary with ground-based surveillance. Compared with standard ground-based surveillance, there are four distinct advantages to observation from the air:
\begin{itemize}
    \item \textit{Scale:} high resolution imaging sensors mounted on airborne platforms enable covering a wide area of observation from the air at multiple resolutions (depending on the altitude  of the aerial platform) with less occlusion \cite{UAV-Human,TinyPersons}. 
    \item \textit{Mobility:} airborne platforms such as UAVs and drones can move rapidly to a target. Once at the destination, they can circle or hover and adjust positions over the target destination for optimum viewing \cite{PoliceDrone,MaritimeSurveillance}. 
    \item \textit{Deployment:} airborne platforms can be deployed at any time (day and night), for any terrain (land and sea), and launched from a long distance \cite{BorderiSurveillance2,MaritimeSurveillance,AVI}.
    \item \textit{Covert observation:} airborne platforms can adjust the flying altitudes to either quietly hide themselves, stay out of visibility reach or explicitly reveal their presence \cite{AEEITS,EyesInTheSky}.
\end{itemize}
However aerial surveillance opens up a plethora of challenges that must be addressed. Aerial surveillance not only inherits all challenges of unconstrained and outdoor surveillance, but also exhibits additional unique challenges to be addressed. From a data perspective, there are seven key challenges to contend with:
\begin{itemize}
    \item \textit{Small resolutions:} objects may appear extremely small in aerial data due to the high flying altitude \cite{TinyPersons,TinyVIRAT}.
    \item \textit{Multiple scales:} multiple instances of  the same class, \emph{e.g.} person, can appear drastically different in sizes and scales \cite{SAMR,ScaleInvarianceAerial}.
    \item \textit{Extreme views:} objects can appear in overhead views, \emph{i.e. }top views and angle views, which rarely exist in generic object detection \cite{GLSA,Drone-Action}.
    \item \textit{Moving cameras:} the view of objects may continuously change due to the moving of the camera mounted on the airborne platform \cite{OkutamaAction,UAV-Human}. This also adds additional challenge of motion blur and camera stabilization \cite{VideoStabilization,AerialMotionBlur}. 
    \item \textit{Non-uniform distribution:} in many cases, objects are distributed non-uniformly: clustered with high density, \emph{e.g.} in a busy urban village \cite{ClusterDet,ClusterNet}; or cluttered with low density in a wide area, \emph{e.g.} in a search and rescue task \cite{SCRDet,NDFT}. 
    \item \textit{Illumination:} non-linear local strong and/or low illumination and lighting due to a wide area coverage \cite{VisDrones}.
    \item \textit{Noise:} the scene may be obstructed, \emph{e.g.} cloud, fog, haze and rain \cite{DomainLabels,NDFT};  The wind may add more challenge in video stabilization for sharp footage. 
\end{itemize}

% https://www.flightglobal.com/military-uavs/could-long-endurance-uavs-be-repurposed-for-surveillance-of-russia-and-china/138823.article

\begin{figure*}
    \centering
    \includegraphics[width=2\columnwidth]{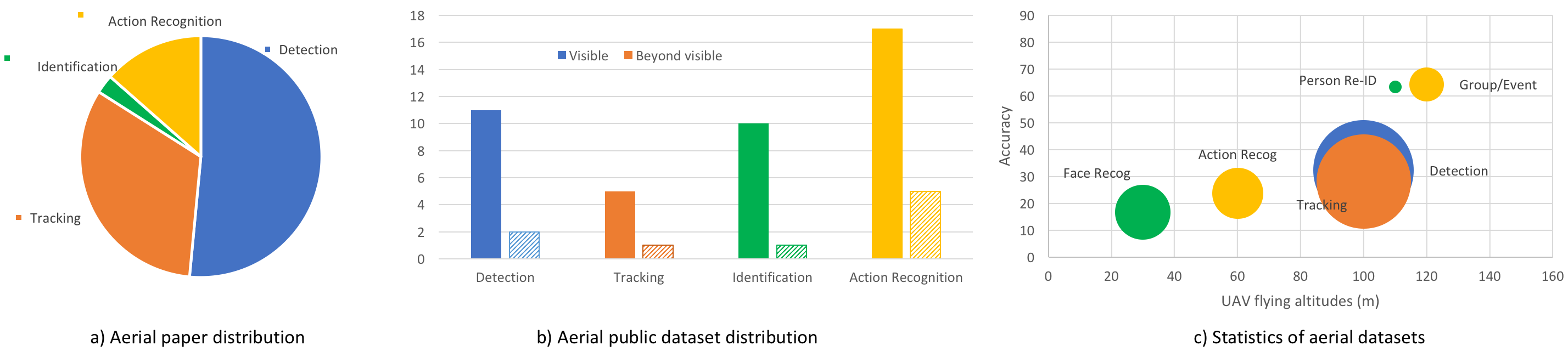}
    \caption{Overview of the state of aerial surveillance tasks. a) The distribution of academic papers across four aerial surveillance tasks: detection, tracking, identification and action recognition; b) The distribution of public aerial datasets across four aerial surveillance tasks: detection, tracking, identification and action recognition. Solid columns denote datasets in the visible spectrum, stripe columns denote datasets in other spectrum; c) The statistics of aerial datasets for aerial surveillance tasks. The size of circles denotes the scale of datasets (\emph{i.e. \#images}, the horizontal axis denotes UAV flying altitudes, and the vertical axis denotes the state-of-the-art accuracies on these datasets.}
    \label{fig:AerialLandscape}
\end{figure*}

\vspace{-6px}
\subsection{State of aerial surveillance}
We provide the first overview of the state of aerial surveillance tasks via analyzing public datasets available for four key tasks, \emph{i.e.} detection, tracking, identification and action recognition, and by summarizing the public interest in this area as illustrated in Fig.~\ref{fig:AerialLandscape}. While this does not reflect efforts by military and defense in these areas, which are usually protected or classified, the state of the public dataset collections is expected to represent a driving factor for the academic research community towards new aerial surveillance tasks.

\begin{itemize}
    \item \textit{Publication distribution:} Fig.~\ref{fig:AerialLandscape}\textcolor{red}{.a} shows that a majority of aerial surveillance papers that are published in open literature focus on detection and tracking, while the number of papers on identification is limited. This would be due to the fact that compared to other tasks, aerial human identification requires high quality visual details of subjects from expensive and purchase-regulated cameras/sensors, which may not be available or too costly to acquire for public use.
    \item \textit{Dataset distribution:} Fig.~\ref{fig:AerialLandscape}\textcolor{red}{.b} shows that a majority of public datasets are for research in detection and action recognition, while the number of datasets for identification is limited. Recent datasets have gone beyond visible cameras, capturing footage from other spectral bands such as thermal, night vision infrared and depth.
    \item \textit{Dataset characteristics:} Fig.~\ref{fig:AerialLandscape}\textcolor{red}{.c} shows that: (i) Flying altitudes: all of the datasets are captured while flying drones or UAVs at most 120 meters, which is the regulated maximum altitude for civilian drones; (ii) Scale: datasets for aerial human detection have large scales, while others usually have small scales; (iii) State-of-the-art accuracies: aerial group/event recognition tasks have already achieved high accuracies in available databases, while face recognition struggles to perform well in the aerial setting.  
\end{itemize}
The following sections (Sections~\ref{sec:aerialobjectdetection} -~\ref{sec:aerialactionrecognition}) dive deep into the four central aerial surveillance tasks.

\begin{comment}

\begin{figure}
    \centering
    \includegraphics[width=0.8\columnwidth]{images/AerialPapers_distribution.png}
    \caption{The distribution of academic papers across four aerial surveillance tasks: detection, tracking, identification and action recognition.}
    \label{fig:AerialPapers_distribution}
\end{figure}

\begin{figure}
    \centering
    \includegraphics[width=0.9\columnwidth]{images/AerialDatasets_distribution1.png}
    \caption{The distribution of public aerial datasets across four aerial surveillance tasks: detection, tracking, identification and action recognition. Solid columns denote datasets in the visible spectrum, stripe columns denote datasets in other spectrum.}
    \label{fig:AerialDatasets_distribution}
\end{figure}

\begin{figure}
    \centering
    \includegraphics[width=\columnwidth]{images/AerialTasks_Overview.png}
    \caption{The statistics of aerial datasets for aerial surveillance tasks. The size of circles denotes the scale of datasets, the horizontal axis denotes UAV flying altitudes, and the vertical axis denotes the state-of-the-art accuracies on these datasets.}
    \label{fig:AerialTasks_Overview}
\end{figure}

\end{comment}

%----------------------------------------------------------------
%\newpage
\section{Aerial Human Detection}
\label{sec:aerialobjectdetection}

%https://github.com/murari023/awesome-aerial-object-detection
Object detection aims at detecting and localizing instances of visual objects of a certain class, \emph{i.e.} human in our surveillance setting, in digital images and videos. 
%Object detection seeks to answer the question: are there any object instances and where are they? 
Akin to ground-based surveillance, aerial object detection also plays a fundamental role as a foundation for high-level tasks in aerial surveillance. However, due to the unique characteristics of aerial imaging settings, aerial object detection exhibits new challenges to be addressed.

In this section, we first review generic ground-based object detection landscape to form a foundation for the object detection discussion. Then, we focus on discussing the aerial human detection in terms of challenges, datasets, approaches, and future outlook. 

\subsection{Challenges for aerial human detection}
\label{sec:AerialHumanDetChallenges}
\begin{figure}
    \centering
    \includegraphics[width=0.7\columnwidth]{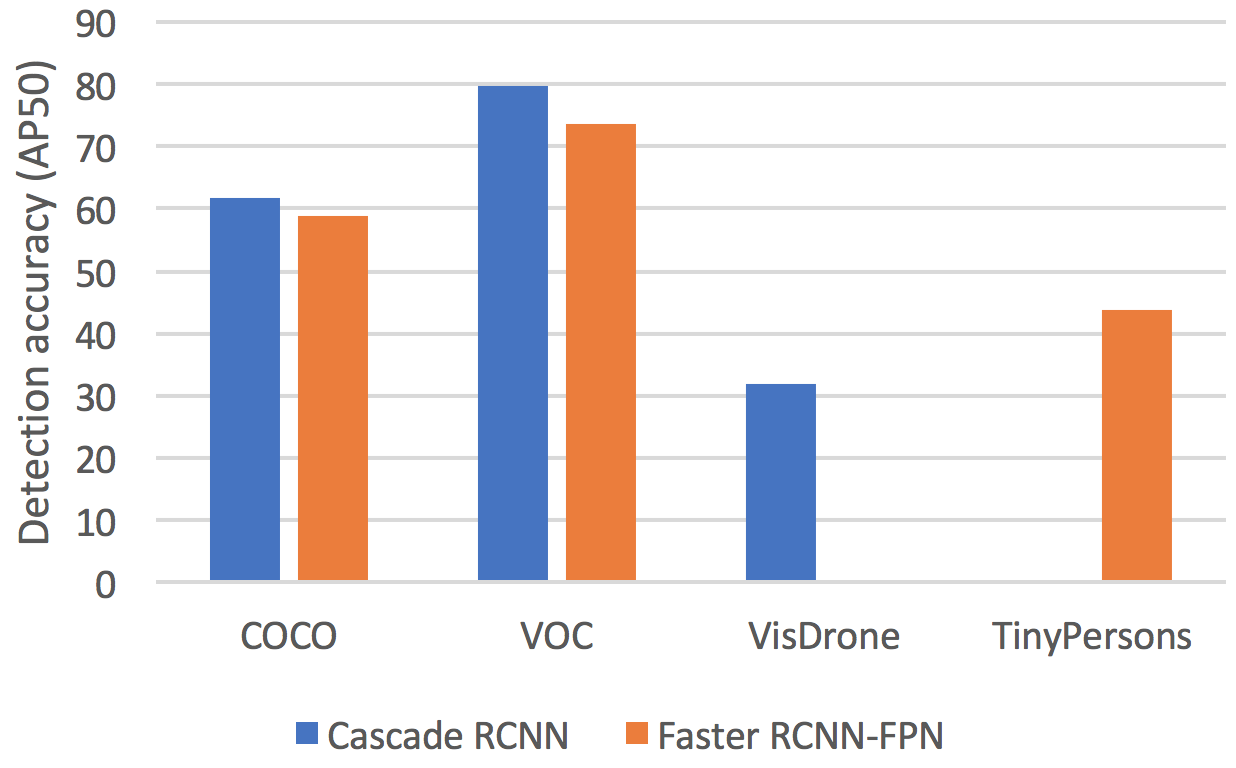}
    \caption{State-of-the-art generic detectors drop the performance when shifting to the aerial data. The figure compares the detection accuracy (AP50) of Cascade R-CNN \cite{CascadeRCNN} and Faster R-CNN \cite{FasterRCNN} on two ground-based (COCO and VOC) and two aerial (VisDrone and TinyPersons) datasets.}
    \label{fig:AerialDetectionPerformance}
\end{figure}

Despite the great success of the generic object detection methods trained on ground-to-ground images, a huge performance drop is observed when they are directly applied to images captured by UAVs \cite{VisDroneDET2020,TinyPersonsChallenge}. Examples of the performance drop of state-of-the-art detectors are illustrated in Fig.~\ref{fig:AerialDetectionPerformance}. The Cascade R-CNN \cite{CascadeRCNN} drops the performance by 50\% in the aerial VisDrone \cite{VisDrones} dataset compared with the ground-based COCO and Pascal VOC datasets \cite{VisDroneDET2020}. The Faster R-CNN \cite{FasterRCNN} also drops the performance by 30\% in the aerial TinyPersons \cite{TinyPersons} dataset compared with the ground-based COCO and Pascal VOC datasets \cite{TinyPersonsChallenge}.

\begin{figure*}
    \centering
    \includegraphics[width=2\columnwidth]{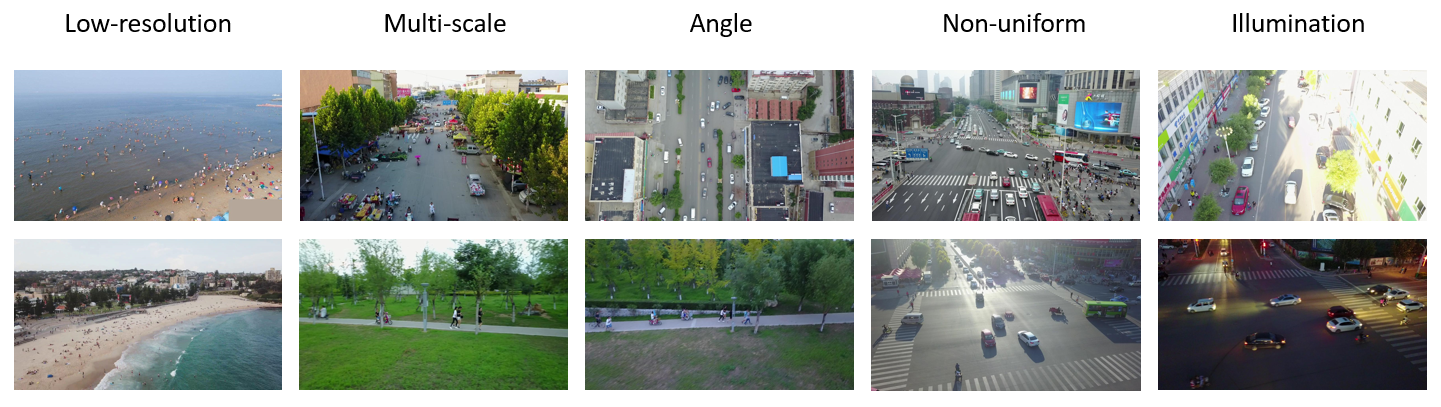}
    \caption{Challenges for aerial object detection: (i) low resolution, (ii) a wide range of scales, (iii) arbitrary viewing angles, (iv) non-uniformly distributed, (v) illumination. Images from the TinyPersons \cite{TinyPersons} and VisDrone \cite{VisDrones} datasets.}
    \label{fig:AerialChallenges}
\end{figure*}

The unsatisfactory performance is owing to the domain shift when compared to ground-based data caused by the high flying altitude and the camera characteristics. Aerial human detection shares all seven challenges discussed in Section~\ref{sec:AdvantagesChallenges}. Examples of these challenges are illustrated in Fig.~\ref{fig:AerialChallenges}.

%While sharing many challenges with unconstrained object detection, aerial object detection has also exhibits novel and pronounced challenges. 

% Cascade R-CNN: COCO VOC vs. VisDrone TinyPerson

\subsection{Datasets for aerial human detection}
The number of aerial datasets has quickly increased in the last few years, partially due to the affordable availability of off-the-shelf drones such as DJI. We summarize public datasets and their statistics in Table~\ref{tab:HumanDetectionDatasets}. Table~\ref{tab:HumanDetectionDatasets} shows a wide range of public datasets where two notable large-scale ones are VisDrone \cite{VisDrones} and TinyPersons 2020  \cite{TinyPersons}. 
\begin{itemize}
  \item \textit{VisDrone 2018-2021} \cite{VisDrones}: The VisDrone team has compiled a dedicated large-scale drone benchmark and organized four consecutive challenges in object detection in ECCV/ICCV from 2018 to 2021. It consists of 400 video clips formed by 265K frames and 10K static images with 2.6M bounding boxes, captured by various drone-mounted cameras, covering a wide range of aspects including location (14 cities), environment (urban and country), objects (pedestrian, vehicles, bicycles, \emph{etc.}), and density (sparse and crowded scenes). %The dataset was collected using various drone platforms (\emph{i.e.} drones with different models), in different scenarios, and under various weather and lighting conditions. 
  \item \textit{TinyPersons 2020}  \cite{TinyPersons}: this dataset is behind two ``Tiny Object Detection'' challenges in ICCV 2019 and ECCV 2020. The unique characteristic of this dataset is the tiny resolution of humans. A majority of human instances appear as small as $[20,32]$ pixels and as tiny as $[2,20]$ pixels. In total, there are 72K objects with bounding boxes that have been manually annotated. %The dataset is split into a training set, which has 794 labeled images with 42K annotations, and a testing set, which has 816 labeled images with 30K annotations.
\end{itemize}
There are a wide range of other datasets with specific characteristics targeting various applications. 
\begin{itemize}
    \item Context: there is a diverse range of contexts, from urban villages \cite{VisDrones,HERIDAL,UAV123}, rural \cite{VisDrones,UAV-Human}, university campuses \cite{StanfordDrones}, internet \cite{TinyPersons}, forest \cite{BIRDSAI}, to agricultural areas \cite{AgriDrone}.
    \item Flying altitudes: there is a diverse range of flying altitudes, from a very close distance of less than 10m \cite{PDESTRE}, to a close range 10-50m \cite{UAV123,StanfordDrones}, to a middle range 50-120m \cite{AgriDrone,UAV-Human}, to above the recreational limit >120m \cite{TinyPersons,MRCNet}. 
    %\item Flying conditions: 
\end{itemize}

\vspace{3px}
\noindent\textbf{Multimodal:} many datasets also provide multi-modal data from other on-board sensors such as GPS, time, altitude, IMU, velocity, weather conditions. The AU-AIR \cite{AUAIR} labeled each frame with time, GPS, IMU, altitude, velocities of the UAV. \cite{NDFT} annotated and released altitude, viewing angle, and weather for the VisDrone dataset. These auxiliary details can be used to improve detection \cite{NDFT}.

\vspace{3px}
\noindent\textbf{Task-specific:} many aerial human detection datasets are aimed for specific tasks. The HERIDAL \cite{HERIDAL,HERIDAL2} and SARD \cite{SARD} datasets were collected to support the search and rescue mission with drones. The AgriDrone \cite{AgriDrone} dataset was collected in an agricultural context for rural applications.

\vspace{3px}
\noindent\textbf{Beyond visible:} sensors from other spectrum have also been employed to complement visible cameras. Infrared sensors complement visible cameras in such adverse conditions as night time or low light. Both the BIRDAI \cite{BIRDSAI} dataset and UAV-Human \cite{UAV-Human} dataset provides thermal-infrared images of humans for detection. Depth sensors complement visible cameras in 3D object details and can be found in the recent UAV-Human \cite{UAV-Human} dataset.

\vspace{3px}
\noindent\textbf{Competitions:} the task of aerial human detection has attracted great attention from the computer vision community. Multiple challenges have been organized in top-tier conferences. The challenges ``Vision meets drones''\cite{VisDroneDET2020} including detection, tracking and crowd counting have been organized in the last four consecutive years in ECCV 2018, ICCV 2019, ECCV 2020 and ICCV 2021 using the VisDrone dataset. The TinyPerson challenges \cite{TinyPersonsChallenge}, organized in ICCV 2019 and ECCV 2020, focus on the persons from a very long distance with a wide view using the TinyPerson dataset. 

Details of these datasets can be found in Appendix~\ref{sec:aerial_humandetection_datasets}.
%Building larger datasets with less bias is critical for developing modern computer vision models in the deep learning era. 

%\tabcolsep=0.5pt
\begin{table}[t]
\small
\renewcommand{\arraystretch}{2}
\caption{Public datasets for aerial human detection. `Spec.', `\#Con', `\#Vid',`\#ID', and `\#Box' respectively represent for the imaging spectrum (V: Visible, T: Thermal-infrared, Z: Depth), the context where data was collected, the number of videos, identities and bounding boxes.}
\label{tab:HumanDetectionDatasets}
\centering
\resizebox{\columnwidth}{!}{
\begin{tabular}{l l l c c r r r r}\toprule
{No.}~~ &  {Dataset} & {Spec.} & {Con.} & {\#Vid.} & {\#ID} & {\#Box.}\\
\midrule
%\midrule
%& Market-1501~\cite{zheng2015scalable}    &Desired  &Real      &6   &1,501  &32,668\\
%& MSMT17~\cite{wei2018person}    &Desired  &Real      &15   &4,101  &126,441\\
\midrule
  {{1}} &  VisDrone~\cite{VisDrones}      & V  & Urban   & 400 & 50 & 2M  \\
  {{2}} & TinyPerson~\cite{TinyPersons}   & V  & Internet & 2 &632 &1,264  \\
  {{3}} & AU-AIR~\cite{AUAIR}             & V  & Real &8 &192 &1,011\\
  {{4}} & MiniDrone~\cite{MiniDrones}     & V  & Carpark & 21 &100 &200\\
  {{5}} & HERIDAL~\cite{HERIDAL}          & V  & Urban &2 &632 &1,264\\
  {{6}} & StanfordDrone~\cite{StanfordDrones}  & V & Campus &2 &450 &900\\
   {{7}} & SARD~\cite{SARD}                & V  & Urban &6 &491 &38,271\\
  {{8}} & UAV123~\cite{UAV123}            & V  & Urban & 123 &632 &1,264\\
  {{9}} & UAVDT~\cite{UAVDT}              & V  & Urban &2 &502 &3,012\\
  {{10}} & PDT-ATV~\cite{DetectionThermal}                & V  & Urban &2 &1,467 &14,000\\
 {{11}} & AgriDrone~\cite{AgriDrone}     & V  & Agri &2 &412 &8,240\\
\midrule
  {{12}} & BIRDSAI~\cite{BIRDSAI}         & VT & Forest &2 &119 & 34K\\
  {{13}} & UAV-Human~\cite{UAV-Human}     & VTZ & Various  & 64K & 1,144 & 41K\\
\bottomrule
\end{tabular}}
\end{table}

\subsection{Approaches for aerial human detection}

%https://github.com/murari023/awesome-aerial-object-detection

%http://www.liuyebin.com/gigavideo/gigavideo.html

Network architecture is the key consideration for all deep-learning-based approaches. The aerial human detection community has adopted two-stage network architecture \cite{DSOD,AdaptiveAnchor,FusionFactor}, one-stage network architecture \cite{DSHNet,FusionFactor,QueryDet}, anchor-free architecture \cite{PENet,RRNet} and network ensembles \cite{VisDroneDET2020,VisDroneDET2020}.

\vspace{3px}
\noindent \textit{Two-stage networks}\\
Due to its performance, two-stage detectors such as Faster R-CNN \cite{FasterRCNN} are popular in the literature \cite{DMNet,PLAOD,GLSA,FusionFactor}. In aerial object detection, a strong multi-scale representation is crucial due to the variance of object sizes and resolutions, which reflects the popularity of the FPN \cite{FPN} backbone in the literature \cite{DSOD,AdaptiveAnchor,FusionFactor}. The multi-scale challenge also makes multi-stage detectors well-suitable for aerial object detection. Many recent approaches \cite{SAMR,SyNet,GDFNet} achieved good performance with Cascade R-CNN \cite{CascadeRCNN}. In fact, many entries, including the winning and runner-up entries, of the VisDrone detection challenges combined Cascade R-CNN \cite{CascadeRCNN} with other networks to achieve high detection performance \cite{VisDroneDET2020}. Similarly, two of top-three detectors in the TinyPersons challenge employed Cascade R-CNN \cite{TinyPersonsChallenge}.

\vspace{3px}
\noindent \textit{One-stage networks}\\
One-stage networks are fast and less computationally intensive than their multi-stage counterparts, hence they have also been employed in many aerial object detection approaches. RetinaNet \cite{RetinaNet} is among the most popular one-stage networks for aerial object detection by \cite{DSHNet,FusionFactor,QueryDet}. EfficientDet \cite{EfficientDet} has also been employed in the aerial setting by \cite{ScaleInvarianceAerial,DomainLabels}. Real-time detectors such as YOLO and its variants \cite{YOLOv4} are also a popular one-stage network for aerial object detection by \cite{GLSA,SelectiveTiling}. Pelee \cite{Pelee}, a real-time object detection system on mobile devices, has also been utilized in aerial object detection by \cite{TilingPower}.

\vspace{3px}
\noindent \textit{Anchor-free}\\
In aerial images where object instances vary drastically in resolution and size in a single image, finding good prior anchor sizes is not feasible. LaLonde \emph{et al.} \cite{ClusterNet} also showed that it is easier to locate extremely small objects with points instead of anchors, hence many approaches have shifted to using anchor-free network designs. CenterNet \cite{CenterNet} with a Hourglass backbone \cite{Hourglass} has been a popular choice such as \cite{AdaptiveSearching,AdaptiveFeat}. \cite{SODA} took advantages of FCOS \cite{FCOS,FCOSPAMI}. Tang \emph{et al.} \cite{PENet} employed a CornerNet design \cite{CornerNet}. Zhang \emph{et al.} \cite{GDFNet} employed a FreeAnchor design \cite{FreeAnchor}. Chen \emph{et al.} \cite{RRNet} also designed a point based detector called RRNet and observed that point-based detectors usually outperform all anchor-based detectors in the VisDrone test set.

\vspace{3px}
\noindent \textit{Network ensemble}\\
Each detector may have different strengths and weaknesses. While the multi-stage detectors tend to produce more false negatives, which means that multi-stage detectors fail to detect some objects, single-stage detectors generally propose more bounding boxes with less quality \cite{SyNet,ObjectDetection20Years}. Hence, combining them may predict more bounding boxes than multi-stage detectors and the quality of the single-stage detector predictions may be enhanced by the multi-stage one. Albaba \emph{et al.} \cite{SyNet} showed that combining a multi-stage detector, \emph{i.e.} Cascade R-CNN \cite{CascadeRCNN}, and a single-stage detector, \emph{i.e.} CenterNet \cite{CenterNet}, yields higher accuracy than individual detectors. The ensemble strategy has been widely employed in practice since it is effective to improve the accuracy of object detection. For example, the winner of the VisDrone object detection challenge 2020 \cite{VisDrones}, DPNetV3 \cite{VisDroneDET2020}, ensembles a few powerful backbones such as HRNet-W40 \cite{HRNet}, Res2Net \cite{Res2Net}, Balanced Feature Pyramid Network \cite{LibraRCNN} and Cascade R-CNN \cite{CascadeRCNN}. The second-ranked detector also uses different combinations of multiple models (\emph{i.e.} Cascade R-CNN \cite{CascadeRCNN}, HRNet \cite{HRNet}, and ATSS \cite{ATSS}) to fuse the detection results.

%\vspace{3px}
%\noindent \textit{Video-based aerial human detection:}

%https://ieeexplore-ieee-org.ezp01.library.qut.edu.au/document/9189139/

%VisDrone-VID2019: The Vision Meets Drone Object Detection in Video Challenge Result
%https://openaccess.thecvf.com/content_ICCVW_2019/papers/VISDrone/Zhu_VisDrone-VID2019_The_Vision_Meets_Drone_Object_Detection_in_Video_Challenge_ICCVW_2019_paper.pdf

\subsection{Techniques to solve aerial detection challenges}
The distinct aerial challenges discussed in Section~\ref{sec:AerialHumanDetChallenges} arise in a large number of domains, across which an effective aerial detection model has to stay robust. 

\vspace{-6px}
\subsubsection{Low resolution and scale variance} 
The most challenging factor in aerial object detection is the low resolution or small size of objects due to the high flying altitude. A human may appear as tiny as a few pixels in an image. For example, the size of humans in the TinyPerson dataset  \cite{TinyPersons} ranges $[2,32]$, in practice $[2,20]$ is considered as tiny and $[20,32]$ is considered as small. In the VisDrone dataset, human sizes only ranges from 0.00014\% to 5.59\% and a mean of 0.044\% and the human objects occupy only three pixels in images within a frame resolution of $1,916\times1,078$ pixels \cite{VisDrones}.
%\cite{ResolutionOnDet,VisDrones}. 
The scale of object resolutions also varies largely from $10^1$ to $10^3$ pixels not just within the dataset, but also within a single image \cite{RRNet}. There are three key approaches to deal with small object detection in aerial data: (i) improve feature maps for small objects \cite{SmallDetSurvey,SmallDetSurvey2}, (ii) incorporate context information of small objects \cite{ContextAerial,ContextRole}, and (iii) data augmentation \cite{VisDroneDET2020}.

%GAN-based small object detection \cite{GANsmallobj,GANsmallobj2}

%Object detection in wide area motion imagery (WAMI)

%https://www.sciencedirect.com/science/article/pii/S0262885620300421

%https://www.sciencedirect.com/science/article/pii/S0957417421000439#b0470

\vspace{-6px}
\subsubsection{Multi-scale detection} 
%Handling multiple scales of objects in images is of crucial importance to object detection \cite{SmallDetSurvey,SmallDetSurvey2}. This becomes even more crucial for aerial object detection due to the dominance of small and multi-scale object instances not only across the whole dataset, but also across a single image. In general object detection, pyramid feature representation strategies such as Feature Pyramid Networks (FPN) \cite{FPN} are the main solution to deal with multi-scale detection. FPN constructed a top-down architecture with lateral connections to produce a series of scale-invariant feature maps, and learned multiple scale-dependent classifiers on these feature pyramids \cite{FPN}. 
Multi-scale detection is commonly used in aerial object detection due to the dominance of small objects and the co-presence of object instances with a wide range of scales. Similar to generic multi-scale object detection, many aerial object detectors \cite{DSOD,PLAOD,USFA,SAMR} employ FPN \cite{FPN} and its variants. New techniques have also been proposed to improve the fusion between feature maps. Gong \emph{et al.} \cite{FusionFactor} introduced a fusion factor to weigh the aggregation between adjacent layers. Gong \emph{et al.} \cite{AdaptiveFeat} also proposed an adaptive feature selection scheme to aggregate feature maps. Wang \emph{et al.} \cite{SAMR} proposed to refine multi-scale features by increasing the receptive field size for high-level semantic features. Differently, Yang \emph{et al.} proposed a sequential approach to employ local high-resolution features for small objects in their QueryDet \cite{QueryDet}. Liu \emph{et al.} proposed to combine both feature pyramid and image pyramid to improve the aggregated features.

\vspace{-3px}
\subsubsection{Viewing angle}
Extreme views due to the high altitude of cameras, birds-eye views, and highly angled views can make objects appear very differently from popular ground-based data. Objects, \emph{i.e.} human, in aerial images may only have a top view and could be arbitrary rotated \cite{ArbitraryOrientedDetection}. If the training data contains these novel views and rotations, most networks can learn to cope with these large variations. However, this requires a very large scale dataset to cover all possible variations. Data augmentation \cite{VisDroneDET2020,TinyPersonsChallenge} and generative models such as GANs \cite{GANdetection} can be used to generate extra data to deal with this challenge during training. 

\vspace{-3px}
\subsubsection{Non-uniform distribution} 
The aerial community has departed from the uniform cropping strategy \cite{TilingPower,SelectiveTiling} by learning density-specific regions to perform detection in parallel with detection on the original image.  Li \emph{et al.} \cite{DMNet} designed a multi-column CNN \cite{MCNN} to predict a density map and utilized a sliding window on the density map to generate proposal regions for cropping. Zhang \emph{et al.} \cite{FullyExploitAerial} designed a Difficult Region Estimation Network to estimate the regions that contain difficult targets and utilized a sliding window on the difficulty map to generate proposal regions for cropping. Yang \emph{et al.} \cite{ClusterDet} designed a Cluster Proposal Network (CPNet) and Tang \emph{et al.} \cite{PENet} designed a coarse network (CPEN) to predict cluster chips which have similar object scales.

%\vspace{3px}
%\noindent \textit{Blurred:}

%\vspace{3px}
%\noindent \textit{Weather:}

\subsection{Techniques to improve aerial human detection}

\vspace{3px}
\noindent \textit{Context modeling:} While feature pyramid networks have the capacity to aggregate multi-scale features to deal with the small and tiny object instances in aerial data, context information can provide strong cues on the presence of humans \cite{ContextRole}. DBNet \cite{VisDroneDET2020}, which ranked the third place in the VisDrone challenge, added a global context block to improve detection. Contextual information can also be captured by dilated convolution \cite{MERI}.

\vspace{3px}
\noindent \textit{Auxiliary meta-data:} Wu \emph{et al.} \cite{NDFT} proposed to incorporate UAV-specific nuisance annotations which are freely available as meta-data in aerial data. For example, for UAVDT \cite{UAVDT} and VisDrones \cite{VisDrones}, three nuisance annotations (altitude, weather, and view angle) are used simultaneously with object classes to train the representation network by a combining a object detection loss and a nuisance prediction loss. Kiefer \emph{et al.} \cite{DomainLabels} proposed a multi-domain strategy to leverage those nuisance annotations. 
%The nuisance annotations are employed to divide images into fine-grained domains. The shared representation network is appended with multiple fine-grained domain heads to train accordingly to the fine-grained domains. For example, for UAVDT \cite{UAVDT} and VisDrones \cite{VisDrones}, three altitude fine-grained domains (low, medium, high), three viewing angle fine-grained domains (front, side, bird) and two lighting fine-grained domains (day, night) are provided. 
The multi-domain strategy improves representation, which subsequently boosts detection accuracy on both datasets.

%\vspace{3px}
%\noindent \textit{Attention modeling} \\

%\vspace{3px}
%\noindent \textit{Bridging the gap between ground-based and aerial data} \\

%GAN

%Overhead data

\subsection{Insights from the competitions}

\begin{table}[]
\caption{Winning entries of the VisDrone-DET2020 aerial object detection challenge (AP - Average Precision) \cite{VisDroneDET2020}.}
\label{tab:VisDrone_2020DET_Leaderboard}
\begin{tabular}{l>{\centering}m{0.2\columnwidth}>{\centering}m{0.2\columnwidth}l>{\centering}m{0.2\columnwidth}}
\hline
\textbf{Method} & \multicolumn{1}{c}{\textbf{AP(\%)}} & \multicolumn{1}{c}{\textbf{AP50(\%)}} & \multicolumn{1}{c}{\textbf{AP75(\%)}} \\ \hline
DPNetV3         & \textbf{37.37}                & \textbf{62.05}                        & \textbf{39.10}                                 \\
SMPNet          & 35.98                               & 59.53                                 & 37.41                                 \\
DBNet           & 35.73                               & 59.63                                 & 36.92                                 \\
ECascade RCNN   & 34.09                               & 56.77                                 & 35.30                                 \\
FPAFS CenterNet & 32.34                               & 56.46                                 & 32.39                                 \\
DOHR RetinaNet  & 21.68                               & 44.59                                 & 18.73                                 \\ \hline
\end{tabular}
\end{table}

\vspace{3px}
\noindent\textit{VisDrone challenge:} results from the VisDrone object detection challenge organized and reported in ECCV 2020 \cite{VisDroneDET2020}, as presented in Table~\ref{tab:VisDrone_2020DET_Leaderboard}, provide two major insights:
\begin{itemize}
    \item The winning entry, DPNetV3, of the challenge achieved an AP50 of 62.05\% \cite{VisDroneDET2020}. This huge increase in performance compared to the generic detectors such as Cascade R-CNN (16.09\%) and approach the accuracy level in the ground-based dataset MSCOCO. 
    \item High-performing entries mainly focus on either (i) directly dealing with aerial challenges, or (ii) combining powers via a network ensemble and data augmentation. For example, DPNetV3 combines Cascade R-CNN with HRNet, Res2Net, FPN and Cascade R-CNN \cite{VisDroneDET2020}. 
\end{itemize}

%\begin{figure}
%    \centering
 %   \caption{Winning entries of the TinyPersons-2020 tiny human detection challenge \cite{VisDroneDET2020}.}
%    \includegraphics[width=\columnwidth]{images/TinyPersons_2020_Leaderboard.png}
%    \label{fig:TinyPersons_Leaderboard}
%\end{figure}

%\vspace{3px}
%\noindent\textit{TinyPersons challenge:} \cite{TinyPersonsChallenge}

\subsection{Aerial human detection beyond visible}
%As near-infrared (NIR) and thermal infrared (TIR) sensors become more affordable in aerial image acquisition, there are increasing interests in detecting humans through 24h all weather scenarios (i.e. fog). There are obvious advantages in NIR and TIR imagery, such as low sensitivities to illumination variations and good capabilities for nightime and bad weather. In addition, different sensory modalities can complement each other in developing a more robust multi-modality human detection framework (\emph{e.g.} visible + infrared). 

%There are two public datatsets with aerial thermal images and annotations available: BIRDSAI \cite{BIRDSAI} and UAV-Human \cite{UAV-Human}. While the BIRDAI dataset \cite{BIRDSAI} purely focuses on thermal images of humans and animals in the forest context, the UAV-Human dataset \cite{UAV-Human} captured multimodal aerial footage simultaneously, including both visible and thermal images. Very few studies \cite{DetectionThermalFCRN,SearchRescureThermal,DetectionThermal} have been conducted regarding this topic. 

Despite the benefits of performing human detection in spectrum other than visible such as 24h all weather scenarios (\emph{i.e.} fog), very few studies have been conducted regarding this topic. Bondi \emph{et al.} showed that Faster-RCNN and YOLO only achieved an accuracy of 18.1\% and 10.4\% respectively on the BIRDSAI \cite{BIRDSAI} dataset. Haider \emph{et al.}\cite{DetectionThermalFCRN}  proposed an convolutional autoencoder architecture to map the human heat signature in the input thermal image to the spatial density maps. Portman \emph{et al.} \cite{DetectionThermal} and Ma \emph{et al.} \cite{DetectionThermal_Handcrafted} showed handcrafted approaches would struggle to deal with aerial thermal human detection due to a large number of variations. Schedl \emph{et al.} showed that aerial thermal person detection under occlusion conditions can be notably improved by combining multi-perspective images before classification \cite{SearchRescureThermal}. Using a synthetic aperture imaging technique, they achieve this with a precision and recall of 96\% and 93\%, respectively on their own dataset.

%\subsectino{Crowd Detection}

%\subsection{Pose Estimation}
%\subsubsection{Generic pose estimation landscape}
%\subsubsection{Aerial pose estimation}
%\noindent\textit{Challenges}\\
%\noindent\textit{Datasets}\\
%\noindent\textit{Approaches}\\
%ACM Graphics: Flycon: Real-time Environment-independent Multi-view Human Pose Estimation with Aerial Vehicles
%\noindent\textit{Gap and future outlook}\\

%\subsection{Aerial face detection}
%\noindent\textit{Challenges}\\
%\noindent\textit{Approaches}\\
%\noindent\textit{Gap and future outlook}\\

%----------------------------------------------------------------
%\newpage
\section{Aerial Human Tracking}
\label{sec:aerialobjecttracking}

%\textcolor{red}{Xiaoming and his postdoc are helping with this section}
%You can refer to Section 3 - Aerial Object Detection as a reference for how we approach the review. The same way to approach will guarantee consistency in the survey. We would also love to hear your suggestion for other sections. 

Object tracking, aiming to estimate the location and scale of an object (\emph{i.e.} pedestrian in our case) in a video with an initial  bounding box given in the first frame, is a key step in video analysis. 
%has been one of the hotspots in computer vision.  
Object tracking is the process of estimating the trajectory of an object in a sequence and consists of four components, \emph{i.e.} object initialization, appearance model, motion prediction and object positioning.
%can be defined as the problem of estimating the trajectory of an object in the image plane among sequential frames. 
To this objective, numerous approaches have been proposed by tackling the following questions: how to model the objects' motion, appearance and shape? which image feature is suitable for tracking? which prior information could be leveraged? The answers to these questions heavily depend on the scenario in which the tracking is performed. For instance, there are many differences between aerial object tracking technology and standard ground object tracking technology due to unique or additional challenges in aerial surveillance settings.

\subsection{Challenges for aerial human tracking}
\begin{figure}
    \centering
    \includegraphics[width=0.95\columnwidth]{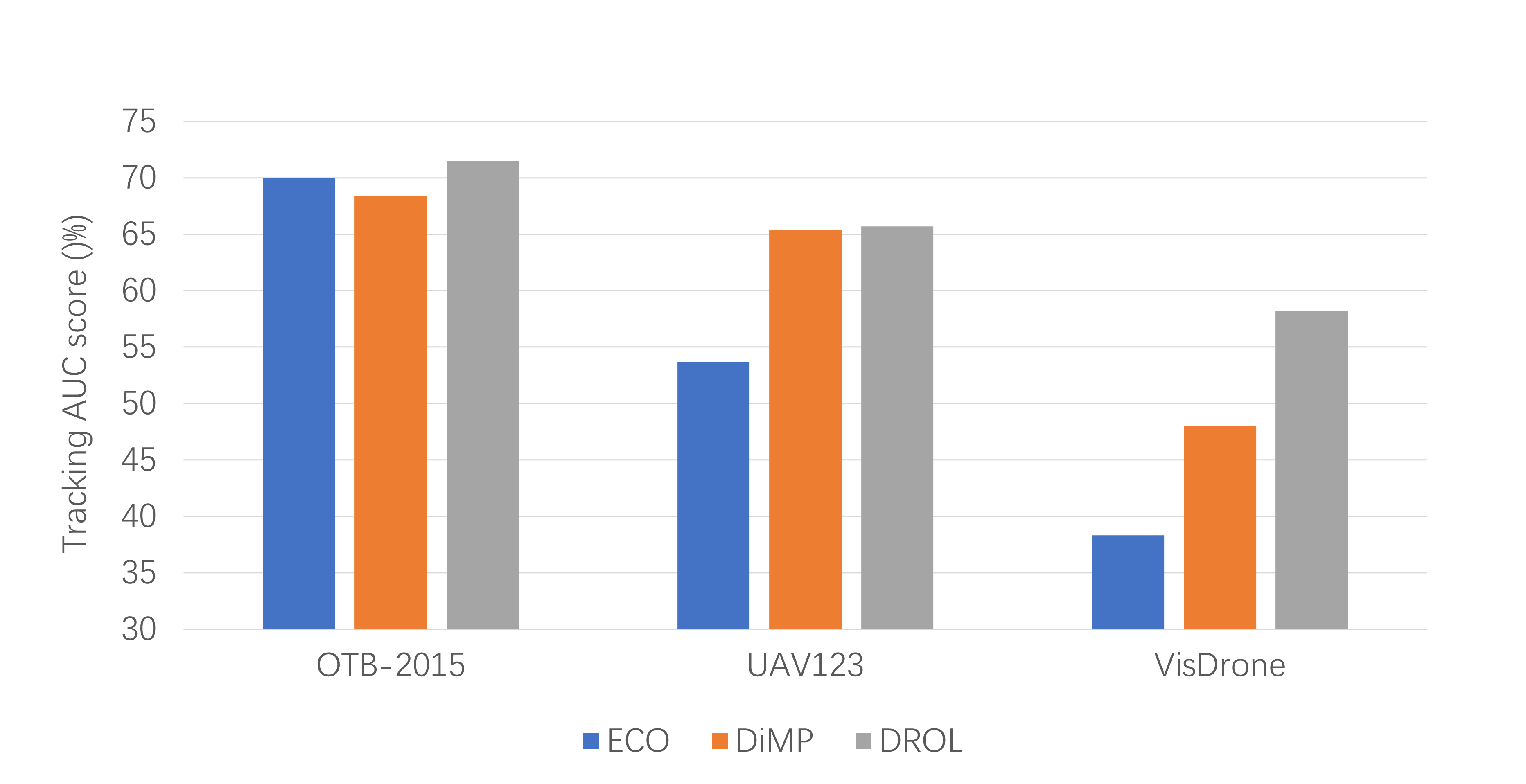}
    \caption{State-of-the-art generic tracking drop the performance when shifting to the aerial data. The figure compares the area-under-the-curve (AUC) score of ECO~\cite{danelljan2017eco}, DiMP~\cite{bhat2019learning} and DROL~\cite{zhou2020discriminative} on one ground-based (DTB-2015~\cite{7001050}) and two aerial (UAV123~\cite{UAV123} and VisDrone~\cite{VisDrones}) datasets.}
    \label{fig:AerialTrackingPerformance}
\end{figure}

Section~\ref{sec:AdvantagesChallenges} and Fig.~\ref{fig:AerialChallenges} detailed the potential challenges in the aerial-based imagery comparing to the ground-based ones. However, there still exists some unique or additional challenges for aerial human tracking. Examples of the performance drop of state-of-the-art trackers are illustrated in Fig.~\ref{fig:AerialTrackingPerformance}.

\begin{itemize}
    \item \textit{Blurred imagery:} Targets for which the image blurred due to the shaking of the UAV fuselage, obstacle avoidance or long imaging distance.
    
    \item \textit{Background clutter:} The background around the target has similar appearances as the target, \emph{i.e.} the building shadows.
    
    \item \textit{Severe occlusion:} The tracked objects may be occluded for a long time, and even disappear for several frames.
    
    \item \textit{Fast-moving targets:} Targets may exhibit fast movement in the image plane. 
    
    \item \textit{Illumination variation:} The illumination of the target changes significantly across frames.  
    
\end{itemize}

\begin{table}[t]
\small
\renewcommand{\arraystretch}{2}
\caption{Public datasets for aerial human tracking. `Spec.', `\#Con.', `\#Vid',`\#ID', and `\#Box.' respectively represent for the imaging spectrum (V: Visible, T: Thermal-infrared, Z: Depth), the context where data was collected, the number of videos, identities and bounding boxes.}
\label{tab:HumanTrackingDatasets}
\centering
\resizebox{\columnwidth}{!}{
\begin{tabular}{l l l c c r r r r}\toprule
 {No.}~~ &  {Dataset} & {Spec.} & {Con.} & {\#Vid.} & {\#ID} & {\#Box.}\\
\midrule
%\midrule
%& Market-1501~\cite{zheng2015scalable}    &Desired  &Real      &6   &1,501  &32,668\\
%& MSMT17~\cite{wei2018person}    &Desired  &Real      &15   &4,101  &126,441\\
\midrule
  {{1}} &  VisDrone~\cite{VisDrones}      & V  & Urban   & 400 & 50 & 2M \\
  {{2}} & UAV123~\cite{UAV123}   & V  & Internet & 123 &632 &1,264 \\
  {{3}} & Campus~\cite{StanfordDrones}             & V  & Real & 8 &192 & 11K\\
  {{4}} & DTB70~\cite{MiniDrones}     & V  & Youtube & 70 &100 &200\\
  {{5}} & UAVDark135~\cite{li2021all}          & V  & Urban &2 &632 &1,264\\
\midrule
  {{6}} & BIRDSAI~\cite{BIRDSAI}         & VT & Forest &2 &119 & 34K\\
\bottomrule
\end{tabular}}
\end{table}

% Cascade R-CNN: COCO VOC vs. VisDrone TinyPerson

\subsection{Datasets for aerial human tracking}
%https://www.mdpi.com/1099-4300/22/12/1358
To date, there is still a limited availability of annotated datasets specific to UAVs where human trackers can be rigorously evaluated or trained for precision and robustness in aerial scenarios.
%there only exist a handful of datasets for aerial human tracking. 
Some drone-captured datasets do not contain either pedestrians or annotations for pedestrians, such as UAVDT dataset~\cite{du2018unmanned}. 
We thus do not list them here.   

We  summarize  public datasets and their statistics in Table~\ref{tab:HumanTrackingDatasets}. There are two notable large-scale datasets: VisDrone \cite{VisDrones} and UAV123 \cite{UAV123}.
\begin{itemize}
    \item \textit{VisDrone 2018-2021:} The VisDrone team has compiled a dedicated large-scale drone benchmark and organized challenges for tracking, \emph{i.e.} VisDrone-SOT2018~\cite{wen2018visdrone} and VisDrone-SOT2019~\cite{du2019visdrone}. The VisDrone-SOT2018 consists of $132$ videos with $106$K frames. Compared with VisDrone-SOT2018, VisDrone-SOT2019 introduces $35$ new sequences. % ($167$ videos with $189$K frames in total). 
    To further increase the diversity of videos and assess the performance of trackers in the wild, VisDrone-SOT2020~\cite{fan2020visdrone} conducts extensive evaluation  of more tracking algorithms using the same dataset in VisDrone-SOT2019. VisDrone2021 further increases the dataset size to $400$ videos with more diverse scenarios.
    \item \textit{UAV123 2021:} this dataset~\cite{UAV123} contains $123$ fully annotated HD sequences over $110$K frames taken from UAV platforms. Each video has $12$ attribute categories.
    %: Aspect Ratio Change (ARC), Background Clutter (BC), Camera Motion (CM), Fast Motion (FM), Full Occlusion (FOC), illumination Variation (IV), Low Resolution (LR), Out-of-View (OV), Partial Occlusion (POC), Similar Object (SOB), Scale Variation (SC), and Viewpoint Change (VC). 
    A video may have a variety of attributes by the shooting conditions. The captured targets include \emph{pedestrian}, vehicles, boats, groups and \emph{etc.} The video resolution is between $720$p and $4$K.
\end{itemize}

\vspace{3px}
\noindent\textbf{Beyond visible:} datasets captured at spectrum other than visible have also been collected. BIRDSAI \cite{BIRDSAI} was collected by a thermal camera mounted on drones. UAVDark 135 \cite{li2021all} targets aerial tracking at night at various scenes such as crossings, t-junctions, roads, and highways. 

Details of these datasets can be found in Appendix~\ref{sec:aerial_tracking_datasets}.

\subsection{Approaches for aerial human tracking}

As mentioned earlier, aerial tracking presents its own challenges compared to generic tracking. 
In aerial imagery, the target appearance changes severely. 
However, most of prevailing aerial trackers are adopted from generic tracking techniques. 
We review the representative approaches based on the following two main streams: discriminative correlation filter (DCF)-based and deep-learning-based.

%AerialMPTNet: Multi-Pedestrian Tracking in Aerial Imagery Using Temporal and Graphical Features

\vspace{3px}
\noindent \textit{Discriminative correlation filter (DCF)-based trackers} \\
    Due to the simplicity and efficiency of the models, DCF-based trackers are widely used for object tracking in aerial videos. 
    Their main advantage is that they can generate plenty of cyclic shift candidates and learn filters in the frequency domain efficiently~\cite{li2020autotrack,MonicaTracking}.
    Wang~\emph{et al.}~\cite{wang2019real} developed a stability measurement metric based on the peak-to-sidelobe ratio, which makes the DCF-based aerial tracker more robust to complicated appearance variations. 
    Huang~\emph{et al.}~\cite{huang2019learning} proposed an aberrance repressed correlation filter, which is capable of suppressing aberrances that is caused by other background noise introduced by conventional DCF and appearance changes of the tracked objects. 
    He~\emph{et al.}~\cite{he2020towards} introduced a unified tri-attention framework to leverage multi-level visual information, including contextual, spatio-temporal and dimension attention to improve UAV tracking robustness and efficiency.
    %the robustness to challenging visual factors such as partial occlusion and clutter background in 
    %
    Ye~\emph{et al.}~\cite{ye2021multi} proposed a novel tracking framework based on a multi-regularized correlation filter, which leads to favorable adaption to object appearance variations and enhancement of discriminability.
    Zhang~\emph{et al.}~\cite{zhang2020object} exploited a two-stage scheme that combines a detection-based network (IoU-Net) with DCF-based tracker for object tracking in aerial videos.
    
\vspace{3px}
\noindent \textit{Deep-learning-based trackers} \\
Even though deep-learning-based trackers have gained tremendous success in ground-view videos, they find it hard to cope with the nuances of aerial videos. 
The main reason is lack of large-scale well-labeled \emph{training} datasets and view-invariant appearance model for fast-moving targets~\cite{song2020cross}. 
To address these challenges,
    CRAC~\cite{song2020cross} introduced a GAN-based tracker to model contextual relation and transfer the relations from ground-view to the aerial-view videos while retaining the discriminative features.
    C2FT~\cite{zhang2018coarse} presented a coarse-to-fine reinforcement learning architecture to address the aspect ratio variation of targets in aerial tracking.
    COMET~\cite{marvasti2020comet} introduced a context-aware IoU-guided tracker that exploits a multitask two-stream network and an offline reference proposal generation strategy to improve the performance of  aerial-view trackers.

%However, a well-designed and robust algorithm is still needed to handle human tracking in aerial setting.

%Feature descriptor is a key component in aerial human tracking. 

%unique sets of challenges than generic human tracking due to target specificity 

%Cross-View Contextual Relation Transferred Network for Unsupervised VehicleTracking in Drone Videos

%Vision Meets Drones: Past, Present and Future

%Visual Tracking With Multiview Trajectory Prediction

%Learning deep representations for ground-to-aerial geolocalization

%Predicting Ground-Level Scene Layout from Aerial Imagery

%https://www.mdpi.com/1099-4300/22/12/1358

\subsection{Techniques to solve aerial tracking challenges}

%\vspace{-3px}
\subsubsection{Blurred imagery} 
Environmental factors and equipment jitter lead to target blur. 
   Deblurring techniques could be used to mitigate this effect. 
   For example, ~\cite{raj2016adaptive} constructs a nonlinear blurred core with multiple moving components. 
   A blind deconvolution technique that used a piecewise linear model was introduced to estimate the unknown kernels, which provides an efficient solution for deblurring motion blurred images.
   
\vspace{-3px}
\subsubsection{Small targets} 
Tracking small targets (\emph{i.e.} human in our case) involves major difficulties comprising the lack of sufficient target information to distinguish them from the background and a large number of possible locations where the subject many appear. Extracting complementary features by employing additional information such as the context leads to more robust feature representations for small objects. 
    The work of~\cite{marvasti2020comet} introduces a context-aware IoU-guided tracker that fully exploits target-related information by multi-scale feature learning and attention modules.
    
     %~\cite{li2020intermittent}combines correlation filter and context-aware information with a new intermittent context learning strategy.

    %The targets (\emph{i.e.}, human in our case) in aerial imagery often occupy a relatively low percentage of pixels on the image plane due to the long distance shooting.
\vspace{-3px}
\subsubsection{Occluded or fast-moving targets} 
Handling occluded or fast-moving targets is of crucial importance in several applications of aerial tracking. The typical solution is to determine the changes in the aspect ratio since targets encounter with fast motion or occlusion often cause aspect ratio variation~\cite{li2017integrating}.

\subsection{Techniques to improve aerial human tracking}

\vspace{3px}
\noindent \textit{Integration of tracking and detection:} In a long aerial video, the targets may frequently leave and re-enter the view (full occlusion). 
A promising scheme to address this challenges is to incorporate a detection process into tracking for re-detecting the tracked targets~\cite{zhou2020tracking}.

\vspace{3px}
\noindent \textit{Motion modeling:} Motion information is crucial for distinguishing tracked targets from clutter background. Motion (including camera and target motion) modeling~\cite{yang2019volumetric} is thus necessary in the aerial tracking. 

\vspace{3px}
\noindent \textit{Domain adaptation:} GANs have been exploited in the context of unsupervised domain adaptation~\cite{volpi2018adversarial}. To alleviate the lack of large-scale well-labeled human videos in aerial tracking, one may consider to transfer the data or feature with diverse variations from ground-view to aerial-view via domain adaptation techniques. For instance,~\cite{song2020cross} proposes a robust tracker with a dual GAN learning mechanism, which can model contextual relation and transfer the ground-view features to the aerial-view ones.

\subsection{Insights from the competitions}
We review the state-of-the-art aerial tracking methods based on their superior performances on 
\emph{VisDrone single object tracking $2020$ challenge}. 
Based on the observation, there are several potential directions that warrant future research. %seem profitable to explore.

\begin{itemize}

   \item \textit{Data augmentation:} Data augmentation is an essential part in network training with limited training data. The augmentation operators for aerial tracking methods include: resale, horizontal flop, rotation, shift, image contrast by Gamma correction and Laplacian operator.
   
   \item \textit{Searching region:} Since the fast-moving and occlusion often occur in aerial setting, the tracking would benefit from a large search region. 
   
    \item \textit{Spatio-temporal context:} the spatio-temporal context information takes an important role for improving the robustness of the trackers. The trackers based on RNN or $3$D-CNN, which leverages the spatio-temporal information, are more effective in coping with the target appearance variations across frames.

\end{itemize}

\subsection{Aerial human tracking beyond visible}

As near-infrared (NIR) and thermal infrared (TIR) sensors become more affordable in aerial image acquisition, there are increasing interests in tracking human through $24$h all weather scenarios (\emph{i.e.} fog). There are obvious advantages in NIR and TIR imagery, such as low sensitivities to illumination variations and good capabilities for nightime and bad weather. 
In addition, different sensory modalities can complement each other in developing a more robust multi-modality tracking framework (\emph{e.g.} visible + infrared).
However, the challenges of background clutter, small target size and occlusion still remain in infrared tracking. 
Besides, public datasets for infrared aerial tracking, not surprisingly, are very scarce. Thus, few studies~\cite{ma2016pedestrian,cao2016two} have been conducted regarding this topic.

\section{Aerial Human Identification}
\label{sec:aerialObjectIdentification}
Beyond detection and tracking, recognizing identity of the object, \emph{i.e. human}, is of paramount importance to aerial surveillance. %Object identification deals with the identity recognition task for instances of visual objects, i.e. human in our surveillance setting, in digital images and videos. Similar to generic ground-based object identification, aerial object identification answers the questions: who is the person detected? Has this person been seen in other cameras? Can we search for all males wearing glasses and red t-shirt from all video feeds? However, due to the unique characteristics of aerial footage,  aerial object identification exhibits unique challenges to be addressed.
To identify humans, biological biometric traits, \emph{e.g.} face, periocular, iris, fingerprint, or behavioral biometric traits, \emph{e.g.} gait, keystroke, voice, signature, cognitive, have been investigated. 
%For aerial human identification, facial and gait are the most feasible clues due to their visibility. 
Aerial human identification is emerging quickly as an important area of research as evidenced by the recent call (in 2021) for Biometric Recognition and Identification at Altitude and Range (BRIAR) from The Intelligence Advanced Research Projects Activity (IARPA) \cite{BRIAR}. The BRIAR program aims to identify or recognize individuals at long-range (\emph{e.g.} 300+ meters), through atmospheric turbulence, or from elevated and/or aerial sensor platforms (\emph{e.g.} $>20^o$ sensor view angle from watch towers or UAV). However, due to the unique characteristics of aerial footage, human identification from aerial footage is very challenging, even for humans \cite{HumanOnAerial,HumanOnAerial2}.

This section reviews face and gait recognition, and person re-identification in the aerial surveillance setting. 

% Check this out: https://www.iarpa.gov/images/files/rfi/IARPA-RFI-19-10.pdf

%%%%%% MENTION BIOMETRIC RECOGNITION AT HIGH ALTITUDE AND LONG RANGE somewhere

%----------------------------------------------------------------
\subsection{Aerial Face Recognition}
\label{sec:facerecog}
Aerial face recognition has sparked enormous interest in recent years, from both positive and negative perspectives. The prospect that police or army could use drones equipped with facial recognition technology to monitor and recognize the identity of each individual in a protest has raised radical concerns about privacy \cite{PoliceDrone}. While facial recognition technology of military and polices is usually kept secret, a multitude of commercial, federal and academic efforts have been made to progress in this area. For example, FA6 Drone claimed to identify a face from a distance of 800 meters at a flying altitude of 100 meters \cite{FaceSix}. Another example is the IARPA initiated benchmark called IJB–S (IARPA Janus Surveillance Video Benchmark) with surveillance videos captured by cameras mounted on a small fixed-wing UAV \cite{IJB-S}. The recent Face In Video Evaluation (FIVE) challenge organized by US National Institute of Science and Technology (NIST) has identification challenges from facial images captured from drones and other aerial vehicles.

\subsubsection{Challenges for aerial face recognition}
%IARPA BRIAR - %https://beta.sam.gov/api/prod/opps/v3/opportunities/resources/files/18f29afa0f894b109257850ac0c3333f/download?api_key=null&token=

%Can Police Face recognition from drones: %https://slate.com/technology/2020/07/police-drone-facial-recognition.html

%Commercial efforts such as FA6 Drone claims to identify a face from a distance of 800 meters at a flying altitude of 100 meters \cite{}.
% https://www.face-six.com/videos/

%If it’s hard to accurately identify faces from a camera on the ground that does not go anywhere, we can assume that it’s even harder to identify faces from pictures taken by a moving drone, where the camera’s view of the person being surveilled is constantly changing

\begin{figure}
    \centering
    \includegraphics[width=0.8\columnwidth]{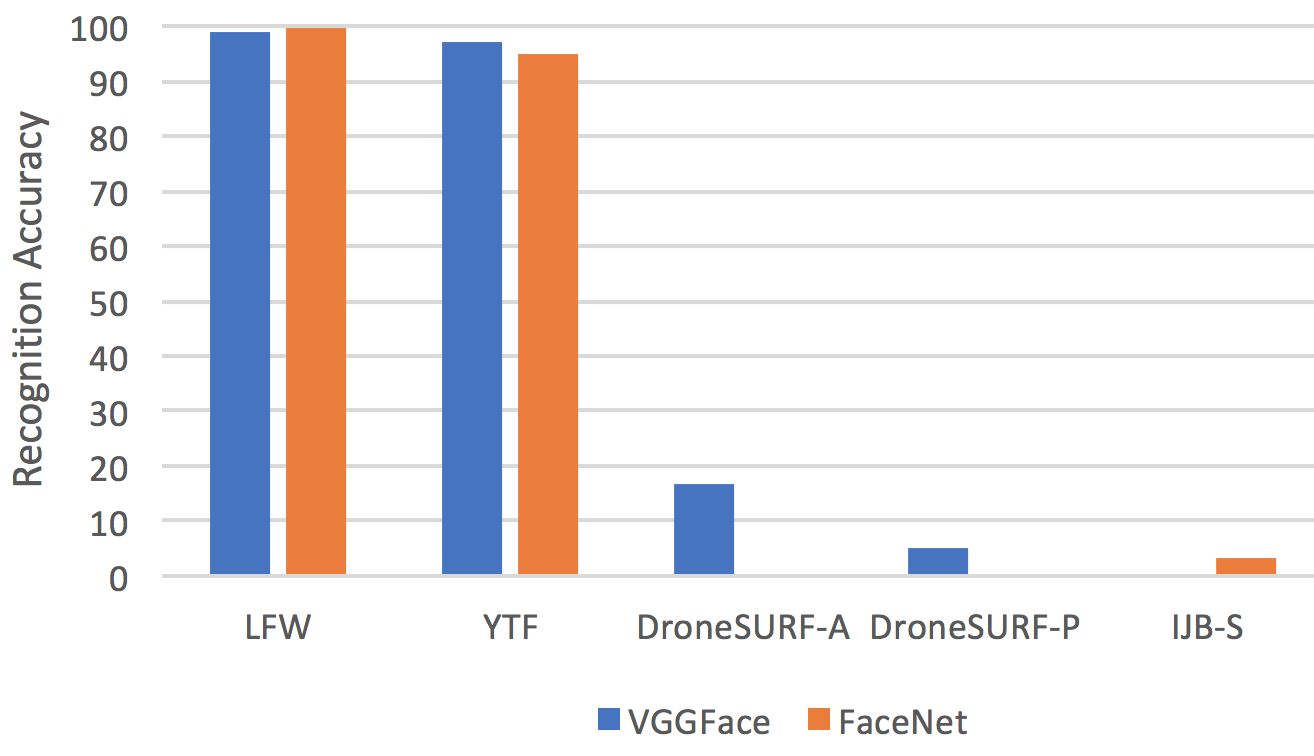}
    \caption{State-of-the-art generic face recognizers drop the performance significantly when shifting to the aerial data. The figure compares the accuracy of VGGFace \cite{VGGFace} and FaceNet \cite{FaceNet} on two ground-based (LFW and YTF) and two aerial (DroneSURF (Active and Passive) and IJB-S) datasets.}
    \label{fig:AerialFRPerformance}
\end{figure}

Despite the great success of the generic face recognition methods trained and tested on ground images, an enormous performance drop is observed when they are directly applied to images captured by UAVs \cite{DroneSURF,E2EunconstrainedFR}. Compared with other tasks, aerial face recognition may exhibit the most drop in performance when migrating from ground to aerial  data. Examples of the performance drop of state-of-the-art face recognizers are illustrated in Fig.~\ref{fig:AerialFRPerformance}. The VGGFace \cite{VGGFace} reduces the accuracy of face recognition from 99.13\% in a ground dataset (LFW) to 16.78\% in an aerial dataset (DroneSURF-Active) where subjects are cooperative and 4.95\% in a dataset (DroneSURF-Passive) where subjects are covertly surveiled \cite{DroneSURF}. The FaceNet \cite{FaceNet} drastically dropped its performance from 99.63\% in a ground dataset (LFW) to 3.33\% in a aerial dataset (IJB-S) \cite{E2EunconstrainedFR}.

The unsatisfactory performance is owing to the domain shift caused by the high flying altitude and the camera characteristics. The small size of a face itself \cite{LRface,SRface} tends to amplify all seven challenges discussed in Section~\ref{sec:AdvantagesChallenges}, making aerial face recognition extremely challenging.

\subsubsection{Datasets for aerial face recognition}
Despite the wide fear of aerial facial recognition with respect to privacy issues, the actual number of public datasets and research in this area is extremely limited. The lack of public datasets may be partially attributed to the privacy concern. There are only three public datasets for this task: DroneFace \cite{DroneFace}, IJB-S-UAV \cite{IJB-S}, and DroneSURF \cite{DroneSURF}. \textbf{DroneFace 2017} \cite{DroneFace} is not real aerial footage where a commercial sport camera, \emph{i.e.} GoPro Hero3+, was set up at a number of fixed altitudes (1.5, 3,4,5m) and distances (2:0.5:17m) to simulate the context that a drone seeks lost people on the streets. \textbf{IJB-S-UAV 2018} \cite{IJB-S} only contains a small number of videos, \emph{i.e.} 10 videos, captured by a fixed-wing UAV flying over a marketplace.

\textbf{DroneSURF 2019} \cite{DroneSURF} is the most realistic dataset for aerial face recognition. It contains 200 videos of 58 subjects, captured across 411K frames, having over 786K face annotations. The dataset demonstrates variations across two surveillance use cases: (i) active and (ii) passive, two locations, and two acquisition times. DroneSURF encapsulates challenges due to the effect of motion, variations in pose, illumination, background, altitude, and resolution, especially due to the large and varying distance between the drone and the subjects. Examples of images from two datasets are shown in Fig.~\ref{fig:Aerial_FR_Datasets}.

Details of these datasets can be found in Appendix~\ref{sec:aerial_facerecognition_datasets}.

\begin{table}[t]
\small
\renewcommand{\arraystretch}{2}
\caption{Public datasets for aerial face recognition. `Con.', `\#ID', `\#Frm' and `\#Vid.' respectively represent the context where data was collected, the number of identities, frames and videos.}
\label{tab:FRDatasets}
\centering
\resizebox{\columnwidth}{!}{
\begin{tabular}{l l l c c r r r r}\toprule
 {No.}~~ &  {Dataset} & {Height (m).} & {Con.} & {\#ID.} & {\#Frm} & {\#Vid.}\\
\midrule
%\midrule
%& Market-1501~\cite{zheng2015scalable}    &Desired  &Real      &6   &1,501  &32,668\\
%& MSMT17~\cite{wei2018person}    &Desired  &Real      &15   &4,101  &126,441\\
\midrule
  {{1}} &  DroneFace~\cite{DroneFace}     & 1-5  & Campus   & 11 & 2,057 & -  \\
  {{2}} & IJB-S~\cite{IJB-S}              & 10  & Marketplace & 202 & 632 & 10  \\
  {{3}} & DroneSURF~\cite{DroneSURF}      & -  & Various & 58 & 411K & 200\\
\bottomrule
\end{tabular}}
\end{table}

%\vspace{3px}
%\noindent\textbf{MUBIDUS-I 2019 \cite{MUBIDUS-I}:}

%\vspace{3px}
%\noindent\textbf{FIVE 2017 \cite{}:} The information about this dataset is not available. 

\begin{figure}
    \centering
    \includegraphics[width=\columnwidth]{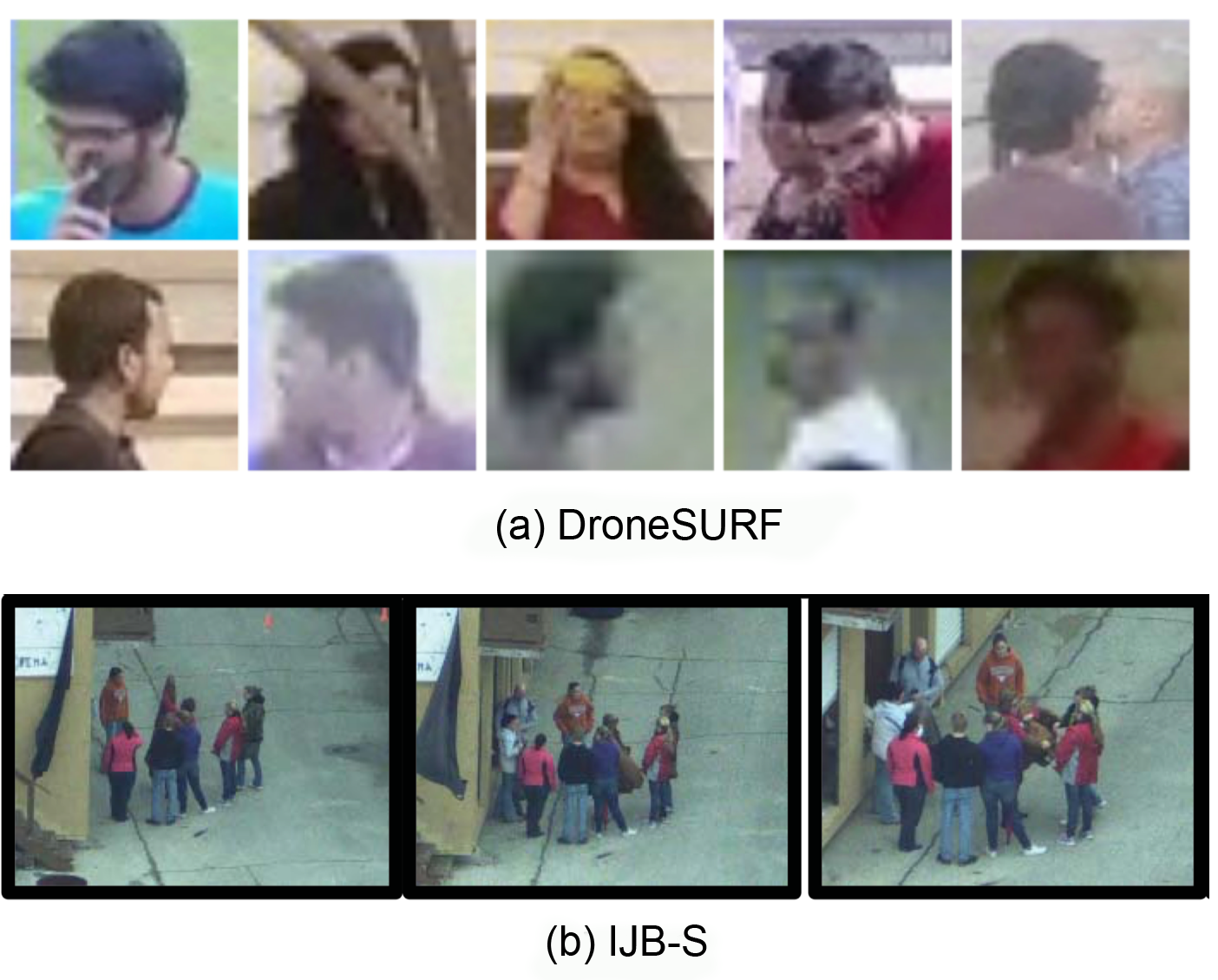}
    \caption{Examples of aerial facial images from two datasets: (a) DroneSURF \cite{DroneSURF} and (b) IJB-S \cite{IJB-S}.}
    \label{fig:Aerial_FR_Datasets}
\end{figure}

\subsubsection{Approaches for aerial face recognition}
Surprisingly the number of published work on aerial face recognition is scarce. There is only one paper on the DroneFace \cite{DBFR}, one paper on the DroneSURF \cite{MRFR}, and four papers on the IJB-S-UAV \cite{E2EunconstrainedFR,CFAN,MARN,REAN}. While targeting aerial facial recognition, all of these had no mechanism to deal with the unique challenges of aerial footage. For example, both \cite{DBFR} and \cite{MRFR} simply employed a classification approach with a cross-entropy loss to classify a probe face into a closed gallery list. All work \cite{CFAN,MARN,REAN} on the IJB-S-UAV simply investigated different strategies to aggregate frames in a video. However, these early attempts have shown the difficulty of aerial facial recognition. For the IJB-S-UAV dataset, the best model \cite{MARN} only achieved a rank-1 accuracy of 7.63\% in a closed-set setting and 3.13\% in an open-set setting. For the DroneSURF dataset, the best model \cite{MRFR} only achieved a recognition accuracy of 24.25\% on the active use case and 3.72\% on the passive use case. When probe images are carefully cropped, the accuracies can reach 60.87\% on the active use case and 45.84\% on the passive use case.

\subsection{Aerial Gait Recognition}
\label{sec:gaitrecog}
%\textcolor{red}{Xiaoming and his postdoc are helping with this section}
%You can refer to Section 3 - Aerial Object Detection as a reference for how we approach the review. The same way to approach will guarantee consistency in the survey. We would also love to hear your suggestion for other sections. 

Gait, the walking pattern of an individual, is one of important behavioral biometric modalities which can operate at a distance without users’ cooperation. Gait recognition has advantages in the aerial scenario, since the dynamic information might be relatively more persistent and  evident at low-resolution imagery than the static biometric characteristics such as the face.

\subsubsection{Challenges for aerial gait recognition}
\begin{figure}
    \centering
    \includegraphics[width=1\columnwidth]{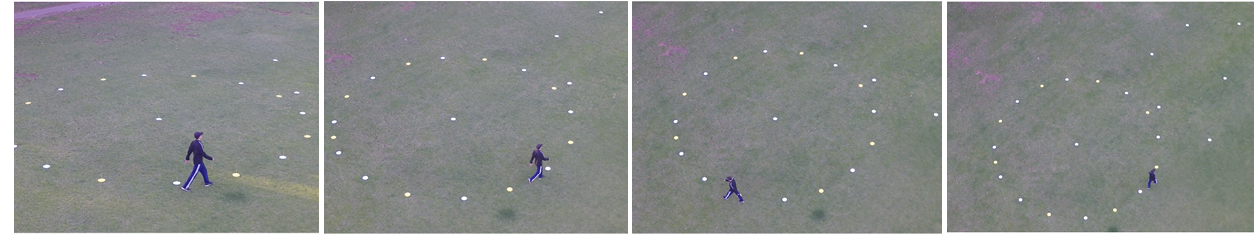}
    \caption{Challenges in aerial gait recognition: large tilt angles and low resolution. Images from the Aerial Gait Dataset~\cite{perera2018human}.}
    \label{fig:Aerial_Gait_Datasets}
\end{figure}

Despite the great progress of generic gait recognition methods trained on ground-to-ground images, the performance of a ground-data-trained model often degrades on aerial data due to the domain shift issue caused by the large discrepancy of pose, resolution and clothing appearance (as shown in Fig.~\ref{fig:Aerial_Gait_Datasets}).

\vspace{3px}
\noindent\textbf{Large pose discrepancy}: Most gallery gaits are captured with ground cameras while the probe gaits are from aerial views with large tilt angles, which makes gait recognition challenging.

\vspace{3px}
\noindent\textbf{Low image resolution}: Given the altitude of the aerial sensor, the probe gait videos are often captured at lower image resolutions, which diminishes the gait information. 

\vspace{3px}
\noindent\textbf{Changing clothing}: This is the classic challenge for gait recognition where the subject in gallery and probe could wear completely different clothing, resulting in large appearance discrepancy. 

%\begin{figure}
%    \centering
%    \includegraphics[width=1\columnwidth]{images/sec_5_2_gait/gait_challenges.png}
%    \caption{Challenges in aerial gait recognition: (i) large tilt angles, (ii) low resolution, (iii) changing clothing. Images from aerial gait dataset~\cite{perera2018human} or FVG dataset~\cite{zhang2019gait}.}
%    \label{fig:Aerial_Gait_Challenge}
%\end{figure}

%Add one figure showing examples of three challenges.

\subsubsection{Datasets for aerial gait recognition}
There is only one aerial gait recognition dataset in the literature: \textbf{Aerial Gait Dataset} \cite{perera2018human}. The gait sequences are captured by a drone with heights ranging from $10m$ to $45m$ while the subject was waling along a circle. All videos in the dataset are in HD format ($1920\times 1080$). However, this dataset is of a small scale with only $17$ video sequences. 
%There are few dataset for the purpose of aerial gait recognition.

%\vspace{3px}
%\noindent\textbf{CASIA-B:} CASIA-B dataset~\cite{yu2006framework} is the most widely used generic gait dataset. It contains $11$ views from $124$ subjects in the form of both RGB and silhouettes under the indoor scenario. It considers $3$ different walking conditions: normal (NM), carrying bag (BG) and wearing a coat (CL), respectively with $6$, $2$ and $2$ gait sequences per person per view.

%\vspace{3px}
%\noindent\textbf{USF:} USF dataset~\cite{sarkar2005humanid} is composed of outdoor $1,870$ sequences from $122$ subjects, who are walking in an elliptical path under a variety of covariates, including carrying a briefcase, walking on different surfaces, wearing different shoes and time elapse.

%\vspace{3px}
%\noindent\textbf{OU-ISIR-LP and OU-ISIR-LP-Bag:} OU-ISIR large population (OU-ISIR-LP) dataset~\cite{iwama2012isir} is a large-scale gait dataset from $4,016$ subjects with ages ranging from $1$ to $94$ years old. The gait sequences have been captured in two different acquisition sessions in indoor halls using  four cameras placed at $55^{\circ}$, $65^{\circ}$, $75^{\circ}$ and $85^{\circ}$. OU-ISIR-LP-Bag dataset~\cite{uddin2018isir} includes gait videos from $62,528$ subjects with various carried objects, \emph{e.g.} handbag, book, umbrella, backpack, and luggage. Example images from four datasets are shown in Fig.~\ref{fig:Aerial_Gait_Datasets}.

%Add one figure showing examples of gait databases.

\subsubsection{Approaches for aerial gait recognition}

There are very few existing gait recognition methods specifically working in the aerial domain.
~\cite{sien2019deep} proposes a two-step framework which first detects human in a video based a Single Shot Multi-box Detection (SSD), and then leverages a LSTM-based recurrent processing structure to perform gait recognition.
~\cite{perera2018human} estimates the gait sequence and movement trajectory of a person from an aerial video. 
The proposed solution consists of perspective correction module, feature extraction module, pose estimation module, and trajectory estimation module. 

\subsection{Aerial Person Re-ID}
\label{sec:PersonReID}
Person Re-ID aims at retrieving a person of interest across multiple non-overlapping cameras. Person Re-ID seeks to answer questions such as ``Where and when has this person been seen in the surveillance network?''. The query person can be represented by visual cues, \emph{i.e} images or videos, or/and textual descriptions \cite{ReIDsurvey}. Similar to ground-based surveillance, aerial person re-ID is usually the most practical human recognition tool in unconstrained surveillance conditions where noisy cameras and uncooperative subjects hinder precise measurements of biometric traits such as the face. However, due to the unique characteristics of aerial footage, aerial person re-ID exhibits unique challenges to be addressed.

\subsubsection{Challenges for aerial person re-ID}
%One of the very early attempts on aerial person re-ID is in 2010 by Oreifej \emph{et al.} \cite{FirstAerialPersonReID} where human instances in each frame are detected by a classic HOG-based SVM, their blobs are extracted and aligned, and the similarity between two human blobs is calculated by the Earth Mover Distance (EMD). The research in this area has grown drastically in both the scale of data and the proposed design to better address the new challenges in the aerial footage. This section presents the unique advantages of challenges of performing person re-ID in aerial settings, the public datasets, the state-of-the-art approaches, and the gap and future outlook for aerial person re-ID. 

\begin{figure}
    \centering
    \includegraphics[width=0.8\columnwidth]{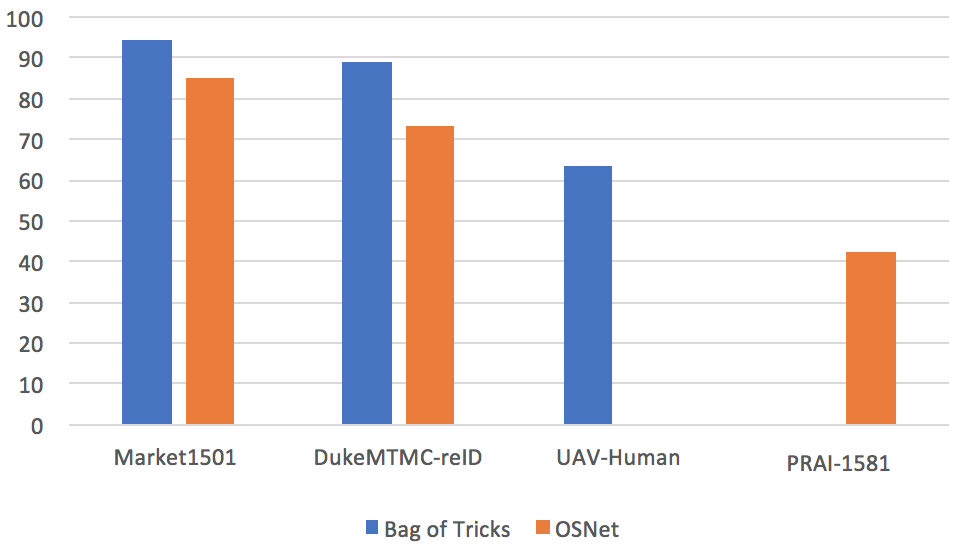}
    \caption{State-of-the-art generic person re-ID algorithms drop the performance when shifting to the aerial data. The figure compares the re-identification accuracy of Bag of Tricks \cite{BoT} and OSNet \cite{OSNet} on two ground-based (Market1501 and DukeMTMC) and two aerial (UAV-Human and PRAI-1581) datasets.}
    \label{fig:AerialReIDPerformanceDrop}
\end{figure}

%Despite the great success of the generic person re-ID methods trained on ground-to-ground images, a huge performance drop is observed when they are directly applied to images captured by UAVs \cite{PRAI1581,UAV-Human}. 
Compared with other tasks, person re-id may exhibit the least drop in performance when migrating from ground to aerial data. Examples of the  performance  drop  of  state-of-the-art person re-id algorithms are illustrated in Fig.~\ref{fig:AerialReIDPerformanceDrop}. The state-of-the-art Bag of Tricks \cite{BoT} reduces the accuracy from 94.2\% and 89.1\% in two ground datasets, Market1501 and DukeMTMC, to 63.4\% in the aerial UAV-Human dataset \cite{UAV-Human}. Similarly, the OSNet \cite{OSNet} reduces the accuracy from 84.9\% and 73.5\% in two ground datasets, Market1501 and DukeMTMC, to 42.1\% in the aerial PRAI-1581 dataset \cite{PRAI1581}.

\begin{figure}
    \centering
    \includegraphics[width=\columnwidth]{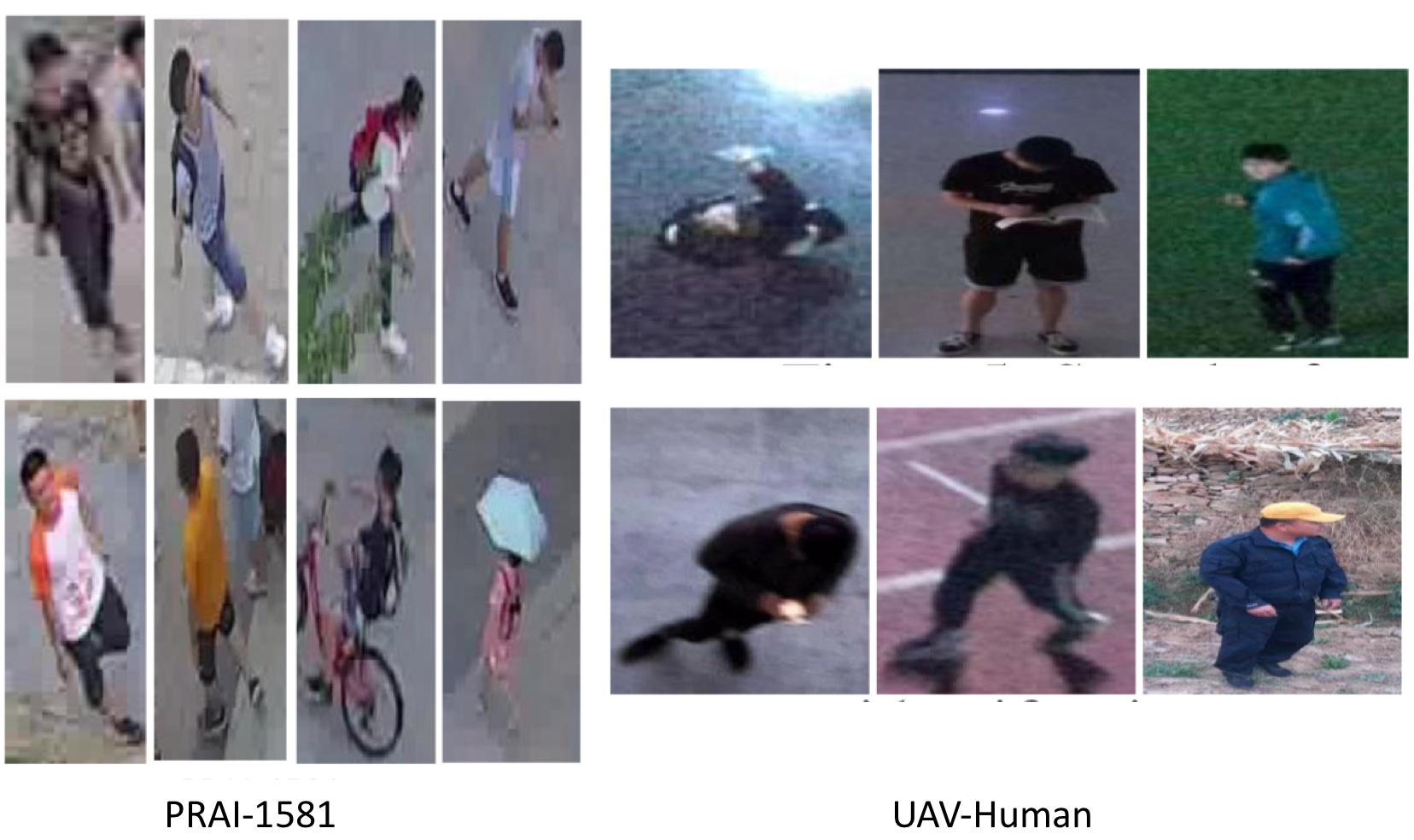}
    \caption{Challenges for aerial person re-ID due to  elevation angle, blur, lighting/illumination, and occlusion. Images from two aerial person re-ID datasets: PRAI-1581 \cite{PRAI1581} and UAV-Human \cite{UAV-Human}.}
    \label{fig:AerialPersonReID_Datasets}
\end{figure}

The performance gap is owing to the domain shift caused by the high flying altitude and the camera characteristics. Aerial Person Re-ID exhibits all seven challenges discussed in Section~\ref{sec:AdvantagesChallenges}. Examples of these challenges are illustrated in Fig.~\ref{fig:AerialPersonReID_Datasets}.

\subsubsection{Datasets for aerial person re-ID}
One of the first datasets for aerial person re-ID is MRP which was collected by Layne \emph{et al.} \cite{MRP} in 2014. Since then, due to the affordable availability of off-the-shelf drones with higher resolution cameras, the number of aerial datasets has quickly emerged in the last three year. We summarize public datasets and their statistics in Table~\ref{tab:aerialReIDDatasets}. Two notable large-scale  datasets are PRAI-1581 \cite{PRAI1581} and UAV-Human \cite{UAV-Human}.
\begin{itemize}
    \item \textit{PRAI-1581 2020} \cite{PRAI1581}: Zhang \emph{et al.} recently collected a large dataset for aerial person re-ID. The images were shot by two DJI drones at an altitude ranging from 20 to 60 meters. The dataset consists of 39k images of 1581 unique subjects. The resolution of persons is low, ranging from 30 to 150 pixels. The high flying altitude makes the diversity of views, poses more extreme. 
    \item \textit{UAV-Human 2021} \cite{UAV-Human}: Li \emph{et al.} just published a new dataset for aerial person re-identification in CVPR 2021. They flied UAVs in multiple urban and rural districts in both daytime and nighttime over three months, hence covering extensive diversities w.r.t. subjects, backgrounds, illuminations, weathers, occlusions, camera motions, and flying altitudes. It contains videos and annotations for multiple tasks, including person re-id and attribute recognition. There are 41,290 frames and 1,144 identities for person re-id and 22,263 frames for attribute recognition. The unique characteristic of the dataset is that multimodal aerial data was captured using a depth sensor (Azure DK), a fisheye camera, and a night-vision camera.
\end{itemize}

\vspace{3px}
\noindent\textbf{Long-term Aerial Re-ID:} two datasets, MEVA \cite{MEVA} and p-DESTRE \cite{PDESTRE} provide footage captured over multiple days where the actors may change in clothing and accessories.

\vspace{3px}
\noindent\textbf{Hetero Aerial Re-ID IR:} two datasets, UAV-Human \cite{UAV-Human} and MEVA \cite{MEVA} provide both RGB and thermal IR footage that can be used for aerial person re-ID across RGB and IR spectrum.

\vspace{3px}
\noindent\textbf{Aerial Human Attribute Recognition:} UAV-Human \cite{UAV-Human} provides 7 human attributes (gender, hat, backpack, upper clothing color and style, as well as lower clothing color
and style), and p-DESTRE \cite{PDESTRE} provides 16 human attributes (gender, age, height, body volume, ethnicity, hair colour, hairstyle, beard, moustache, glasses, head accessories’, ‘body accessories’, ‘action’ and ‘clothing information’ (x3)). These attributes can be used for aerial attribute recognition, person search or to support the aerial re-ID task.

Details of these datasets can be found in Appendix~\ref{sec:aerial_personreid_datasets}.

\begin{table}[t]
\small
\renewcommand{\arraystretch}{2}
\caption{Public datasets for aerial person re-ID and search. `Height.', `\#Con.', `\#ID' and `\#Frm.' respectively represent the altitude, the context where data was collected, the number of subjects, and frames.}
\label{tab:aerialReIDDatasets}
\centering
\resizebox{\columnwidth}{!}{
\begin{tabular}{l l l c c r r r r}\toprule
 {No.}~~ &  {Dataset} & {Height.} & {Con.} & {\#ID.} & {\#Frm} & {\#Vid.}\\
\midrule
%\midrule
%& Market-1501~\cite{zheng2015scalable}    &Desired  &Real      &6   &1,501  &32,668\\
%& MSMT17~\cite{wei2018person}    &Desired  &Real      &15   &4,101  &126,441\\
\midrule
  {{1}} &  MRP 2014~\cite{MRP}       & 10  & Campus   & 28 & 4,096 & -  \\
  {{2}} & DroneHIT~\cite{IJB-S}      & 25  & Campus & 101 & 46K & -  \\
  {{3}} & p-DESTRE~\cite{PDESTRE}    & 5-6  & Campus & 269 & 14.8M & -\\
  {{4}} & PRAI-1581~\cite{PRAI1581}  & 20-60  & Various & 1,581 & 39K & -\\
  {{5}} & UAV-Human~\cite{UAV-Human} & 10  & Various & 1,144 & 41K & -\\
  {{6}} & MEVA~\cite{MEVA}           & 10  & Various & 100 & - & -\\
\bottomrule
\end{tabular}}
\end{table}

\subsubsection{Approaches for aerial person re-ID}

\vspace{3px}
\noindent\textbf{Closed-world Aerial Person Re-ID} \\
Closed world person re-ID, refers to the search where the  person in the probe image is definitely present as one of the candidates in the gallery images. The majority of prevailing approaches for closed world person re-ID focus on domain invariant settings, \emph{i.e.} ground-ground and aerial-aerial person re-ID. Classical handcrafted features and classical distance metrics have been prevalent in the early stages of the development of closed world person re-ID techniques.

\vspace{3px}
\noindent\textit{Handcrafted feature representation:} Before the explosion of deep learning in 2012, handcrafted features had been employed for aerial person re-ID. Oreifej \emph{et al.} \cite{FirstAerialPersonReID} directly estimated the similarity between two human blobs by the Earth Mover Distance (EMD). Layne \emph{et al.} \cite{MRP} classified human detections into a pre-defined ID list using a number of SVM variants. Schumann \emph{et al.} \cite{AerialPersonReID_CovFu} employed a covariance descriptor \cite{ReID_Cov} and a geodesic distance between two covariance descriptors to measure their similarity. 

\vspace{3px}
\noindent\textit{Learnable feature representation:} All modern approaches have employed CNNs to learn representation directly from data. There are two categories of applying CNNs in person re-ID: discriminative and pairwise. Discriminative approaches classify each human detections into a pre-defined ID list. Schumann \emph{et al.} \cite{ReID_Cov,AerialPersonReID_Attri} designed their own CNN architecture with Inception \cite{InceptionNet} and Residual layers \cite{ResNet} to learn representation. One of the branch in the Grigorev \emph{et al.}'s framework \cite{DroneHIT} employed Resnet-50 \cite{ResNet} as a base network. They trained the base network using a large margin Gaussian mixture (L-GM) loss \cite{LGMloss}. Pairwise approaches seek to directly calculate the similarity between two human detections, eliminating the need for a pre-collected gallery list. Most approaches in the literature combine both discriminative losses and pairwise losses to train the backbone network. Grigorev \emph{et al.} \cite{DroneHIT} trained a backbone ResNet by both a triplet loss and a L-GM loss \cite{LGMloss}. Zhang \emph{et al.} also combine a triplet loss with an identification loss to train the deep network. They experimented with multiple backbone networks coupled with a subspace pooling layer to learn a compact representation \cite{PRAI1581}.

\vspace{3px}
\noindent\textit{Data augmentation:} Data augmentation during training can be applied to increase the robustness against such error sources as viewpoint variations and occlusions. Moritz \emph{et al.} proposed two augmentation approaches, random rotation (RR) and random cropped rotation (RCR), to specifically improve the robustness against diverse perspectives in UAV-based person re-id \cite{ReIDdesign}. Testing with the OSNet baseline \cite{OSNet}, these augmentation techniques can boost the accuracy from 50.0\% to 53.0\% in the PRAI-1581 \cite{PRAI1581} dataset and
from 80.1\% to 83.7\% in the p-DESTRE \cite{PDESTRE} dataset.

\vspace{3px}
\noindent\textit{Transfer learning:} Many approaches \cite{AerialPersonReID_CovFu,AerialPersonReID_Attri,DroneHIT} take advantages of a large number of existing datasets in a classic ground-based setting such as  for transfer learning. A person re-ID model is first trained with these ground-based datasets, then either applied directly \cite{AerialPersonReID_CovFu} or fine-tuned \cite{AerialPersonReID_Attri,DroneHIT} on the aerial dataset. While this strategy has shown to be working, it is still an open question to what extent the pre-training can help, since the characteristics of aerial images and ground-based images differ. 

%\vspace{3px}
%\noindent\textit{Video-based aerial person re-ID}

\vspace{3px}
\noindent\textbf{Open-world Aerial Person Re-ID} \\
Open-world person re-ID is a more difficult problem of person re-ID where the person in the probe may or may not be in the gallery. Many approaches for open world person re-ID depart from the closed-world setting and explore aerial person re-ID across multiple modalities. This includes across domains (aerial-ground \cite{AerialPersonReID_CovFu}), across spectrum (visible-infrared \cite{UAV-Human}), and across modalities (aerial-attribute \cite{PDESTRE}). 

\vspace{3px}
\noindent\textit{Long-term aerial person re-ID:} Long-term aerial person re-ID relaxes the time constraints between the probe and gallery imaging moment. Existing re-ID methods rely on appearance features like clothes, shoes, hair, \emph{etc.} Such features, however, can change drastically from one day to the next, leading to inability to identify people over extended time periods. The p-DESTRE dataset \cite{PDESTRE} also provides a baseline for long-term aerial person re-id based on facial and body features. The facial feature representation was obtained using the ArcFace model \cite{ArcFace} and the body feature representation was obtained using the COSAM \cite{COSAM} model. The Euclidean norm was used as distance function between the concatenated representations. This approach achieved an mAP of 34.9\%, but it is noticeable that the p-DESTRE footage was captured with very low flying altitudes, \emph{i.e.} 5-6m.

\vspace{3px}
\noindent\textit{Aerial-Ground person re-ID:} 
Person re-ID between aerial images and ground-based images is of significant interest to surveillance in such tasks as large scale search. However, the differences in human appearances such as views, poses and resolutions make it very challenging. The representation has to be robust to these variations. There exists one work in aerial-ground person re-id by Schumann \emph{et al.} \cite{AerialPersonReID_CovFu} where the authors attempted to encode a robust representation via covariance descriptors. However, the views of the aerial footage and the ground-based footage are similar. In addition, the dataset is very small with only 1217 probe images and 4244 gallery images \cite{AerialPersonReID_CovFu}. The field is lacking large scale datasets of aerial and ground-based footage for deep networks to learn representation robust to the challenges in views, poses and resolutions. In addition, the field is also lacking approaches to explicitly deal with these challenges. Such generative models such as Generative Adversarial Networks (GANs) \cite{GANsurvey} can be used to synthesize new views, poses or resolutions for robust matching. 

\vspace{3px}
\noindent\textit{Aerial-Attribute human recognition:}
Searching for persons in aerial data based on attribute descriptions such as gender, height, clothing, \emph{etc.} is also of practical significance in surveillance, which is imperative when the visual image of query person cannot be obtained. The re-id task now can be compiled as a human attribute recognition problem \cite{HumanAttriSurvey}. Kumar \emph{et al.} \cite{PDESTRE} has addressed this issue by collecting the aerial dataset, p-DESTRE, and annotating persons with 16 attributes. They employed the COSAM algorithm \cite{COSAM} to search for human with facial and body attributes, which can deal with the long-term re-id task.

\section{Aerial Human Action Recognition}
\label{sec:aerialactionrecognition}

\subsection{Challenges for aerial action recognition}
\label{sec:aerial_actionrec_challenges}
\begin{figure}
    \centering
    \includegraphics[width=0.8\columnwidth]{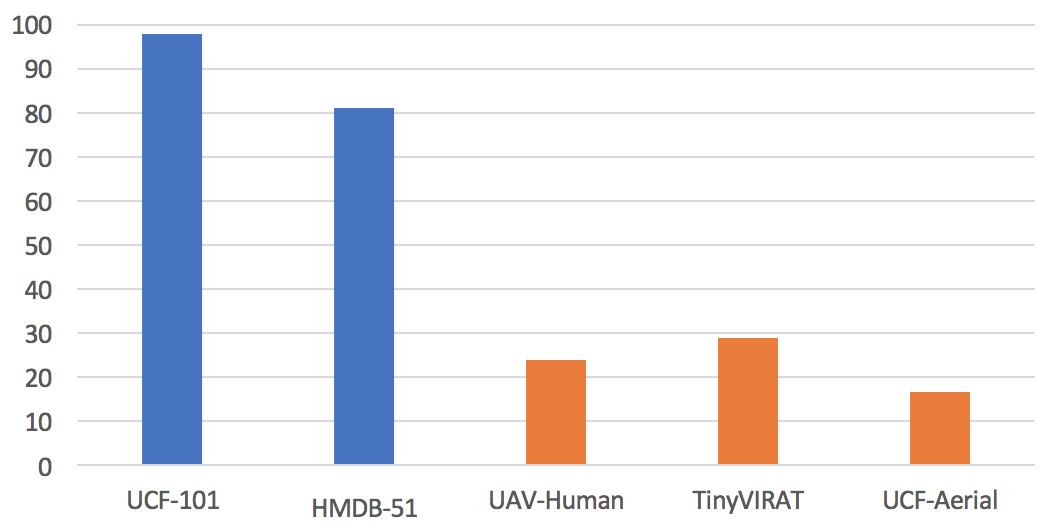}
    \caption{State-of-the-art generic action recognition algorithms drop the performance significantly when shifting to the aerial data. The figure compares the accuracy of I3D \cite{I3D} on two ground-based (UCF-101 and HMDB-51) and three aerial (UAV-Human, TinyVIRAT and UCF-Aerial) datasets.}
    \label{fig:AerialActionPerformanceDrop}
\end{figure}

Despite the great success of the generic human action recognition methods trained on ground-to-ground videos, a huge performance drop is observed when they are directly applied to images captured by UAVs \cite{UAV-Human,MEVA}. Examples  of  the  performance  drop  of the state-of-the-art action recognition algorithm, I3D \cite{I3D} are illustrated in Fig.~\ref{fig:AerialActionPerformanceDrop}. While I3D \cite{I3D} reached 98.0\% and 80.9\% on two popular ground-based action datasets, UCF-101 \cite{UCF101} and HMDB-51 \cite{HMDB51} respectively, it only achieved 23.86\%, 28.7\% and 16.8\% on three aerial action recognition datasets: UAV-Human \cite{UAV-Human}, TinyVIRAT \cite{TinyVIRAT}, and UCF-Aerial \cite{UCF_2011}.

\begin{figure*}
    \centering
    \includegraphics[width=1.6\columnwidth]{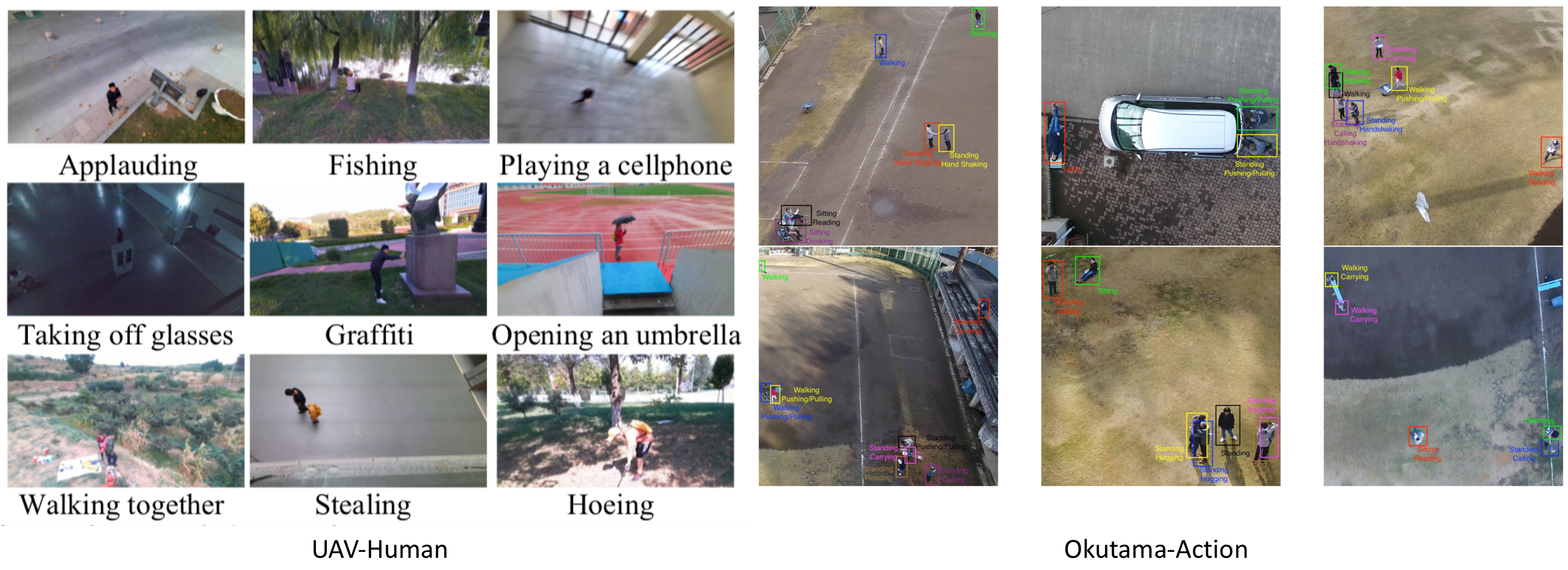}
    \caption{Challenges for aerial action recognition due to the small size of subjects, elevation angle, extreme lighting/illumination, and occlusion. Images from two aerial datasets: UAV-Human \cite{UAV-Human} and Okutama-Action \cite{OkutamaAction}.}
    \label{fig:AerialActionRecog_Datasets}
\end{figure*}

The unsatisfactory performance is owing to the domain shift caused by the high flying altitude and the camera characteristics. Aerial action recognition exhibits all seven challenges discussed in Section~\ref{sec:AdvantagesChallenges}. Especially, the novel elevation views require the action recognition task to explore more semantic attributes to be invariant to the change in views and poses of subjects. The non-uniform distribution requires effective algorithms to explore robust attention mechanism and attend to partial features to be effective. Examples of these challenges are illustrated in Fig.~\ref{fig:AerialActionRecog_Datasets}.

\subsection{Datasets for aerial action recognition}
%The literature reveals four levels of human behaviors as shown in Fig.~\ref{fig:4levelsbehaviors}. The most fine-grained level is gesture recognition, which usually employed for controlling drones/UAVs at close distances \cite{UAV-GESTURE,DDIR}. A majority of aerial datasets focus on the middle two levels, in individual action recognition and human-object or human-human interaction recognition \cite{UAV-Human,UT-Interaction,Drone-Action}. Compared with ground-based surveillance configurations, the aerial configuration is much relevant to group activity understanding and crowd analysis considering the large view of aerial sensors \cite{ERA}. 

%\begin{figure}
%    \centering
%    \includegraphics[width=0.7\columnwidth]{images/HumanBehaviourLevels.jpg}
%    \caption{Four levels of human behaviors \cite{CrowdAnalysis}.}
%    \label{fig:4levelsbehaviors} 
%\end{figure}

%\subsubsection{Aerial Action and Interaction datasets}
A large number of datasets are captured and annotated for aerial action recognition \cite{UAV-Human,Drone-Action,OkutamaAction}. Most popular actions of interest are running, walking, sitting, and standing \cite{UAV-Human,Drone-Action,OkutamaAction}. Among them, many actions are performed in relation to the interaction with objects \cite{OkutamaAction,UAV-Human} and other humans \cite{AVI,UT-Interaction,STPAN}. Popular human-object interactions are reading, carrying, pushing/pulling \cite{OkutamaAction}, fishing and cutting tree \cite{UAV-Human}, entering and exiting a vehicle \cite{MEVA,VIRAT}. Popular human-human interactions can be in forms of violence \cite{AVI,Drone-Action}, social interactions \cite{UAV-Human}. Two most notable aerial action and interaction datasets are UAV-Human \cite{UAV-Human} and MEVA \cite{MEVA}.
\begin{itemize}
    \item \textit{UAV-Human} \cite{UAV-Human}: beside the data and annotations for person re-id and attribute recognition as discussed before, the UAV-Human team has also compiled 67K multi-modal video sequences and 119 subjects for action recognition. The unique characteristic in multimodal aerial data including a depth sensor (Azure DK), a fisheye camera, and a night-vision camera makes this dataset very attractive to the research community.
    \item \textit{MEVA} \cite{MEVA}: this collection observed approximately 100 actors performing scripted scenarios and spontaneous background activity over a three-week period at an access-controlled venue, collecting in multiple modalities with overlapping and non-overlapping indoor and outdoor viewpoints. The resulting data includes videos from 38 RGB and thermal IR cameras, 42 hours of UAV footage, as well as GPS locations for the actors.
\end{itemize}

%Multi-modal:

\vspace{3px}
\noindent\textbf{Task-specific:} many  aerial  human  detection  datasets  are aimed for specific tasks such as sport \cite{NEC-Drone}, search and rescue \cite{DIASR}, synthetic data from game engines \cite{Game-Action}, and multiview \cite{MOD20}. AVI \cite{AVI} is for violent recognition from aerial videos.

\vspace{3px}
\noindent\textbf{Beyond visible:} other sensors have also been employed in parallel with visible cameras. UAV-Human \cite{UAV-Human} employed depth and infrared sensors in parallel to the visible cameras. MEVA \cite{MEVA} employed infrared sensors simultaneously with the visible cameras.

\vspace{3px}
\noindent\textbf{Fine-grained gesture recognition:} some datasets are collected to perform gesture recognition from aerial images/videos, which are usually employed for controlling drones/UAVs, such as UAV-Gesture \cite{UAV-GESTURE} and DDIR \cite{DDIR}. Due to the fine-grained recognition, the images/videos are usually captured at short distances and low flying altitude.

\vspace{3px}
\noindent\textbf{Large scale crowd behaviors:} Compared with ground-based surveillance configurations, the aerial configuration is much relevant to group activity understanding and crowd analysis considering the large view of aerial sensors \cite{ERA}. Datasets such as DroneCrowd \cite{DroneCrowd} and DLR-ACD \cite{MRCNet} focus on crowd counting and density estimation while ERA \cite{ERA} classifies aerial videos into action classes.

%\subsubsection{Aerial Group Activity datasets}

%DroneCrowd dataset CVPR21

Details of these datasets can be found in Appendix~\ref{sec:aerial_humanaction_datasets}.

\begin{table}[t]
\renewcommand{\arraystretch}{2}
\caption{Public datasets for aerial human action recognition. `Spec.', `\#Con.', `\#Vid' and `\#Box.' respectively represent for the action type (A: Action and Interaction, G: Gesture, C: Group event/activity crowd analysis), the context where data was collected, the number of videos, identities and bounding boxes.}
\label{tab:HumanActionDatasets}
\centering
\resizebox{\columnwidth}{!}{
\begin{tabular}{l l l c c r r r r}\toprule
 {No.}~~ &  {Dataset} & {Lev.} & {Con.} & {\#Vid.} & {\#ID} & {\#Hei.}& {\#Act.}\\
\midrule
%\midrule
%& Market-1501~\cite{zheng2015scalable}    &Desired  &Real      &6   &1,501  &32,668\\
%& MSMT17~\cite{wei2018person}    &Desired  &Real      &15   &4,101  &126,441\\
\midrule
  {{1}} &  UAV-Human~\cite{UAV-Human}      & A  & Urban   & 67K & 119 & - & 155\\
  {{2}} & Drone-Action~\cite{Drone-Action}   & A  & YT & 240 & 10 & 8-12 & 13 \\
  {{3}} & Okutama-Action~\cite{OkutamaAction}   & A  & Urban & 43 & 9 & 10-45 & 12 \\
  {{4}} & UCF-ARG~\cite{UCF_2011}     & A  & Carpark & 480 & 12 & - & 10\\
  {{5}} & UCF-Aerial~\cite{UCF_2008}     & A  & Carpark & 480 & 12 & - & 10\\
  {{6}} & MEVA~\cite{MEVA}          & A  & Urban & 4.6h & 100 & - & 37\\
  {{7}} & MOD20~\cite{MOD20}            & A  & YT & 2K & - & - & 20\\
  {{8}} & VIRAT~\cite{VIRAT}            & A  & Carpark & 17 & - & - & 23\\
  {{9}} & NEC-Drone~\cite{NEC-Drone}  & A & Gym & 5K & 19 & - & 16\\
   {{10}} & Game-Action~\cite{Game-Action}    & A  & Game & 200 & - & - & 7\\
  {{11}} & AVI~\cite{AVI}            & A  & Urban & 2K & 25 & 2-8 & 5\\
  {{12}} & DIASR~\cite{DIASR}        & A  & Disaster & 7K & - & 10-40 & \\
  {{13}} & Youtube-Aerial~\cite{Game-Action}    & A  & YT & 400 & - & - & 6\\
  {{14}} & UT-Interaction~\cite{UT-Interaction}  & A & Forest & 60 & 6 & - & 6\\
\midrule
  {{15}} & UAV-Gesture~\cite{UAV-GESTURE}         & G & Forest & 119 & 10 & - & 13\\
  {{16}} & DDIR~\cite{DDIR}     & G & Various  & - & 26 & 2-10 & 9\\
\midrule
  {{17}} & ERA~\cite{ERA}         & C & YT & 2.8K & - & - & 25\\
  {{18}} & DLR-ACD~\cite{MRCNet}         & C & YT & 2.8K & - & - & 25\\
  {{19}} & VisDroneCC~\cite{VisDrones}         & C & YT & 2.8K & - & - & 25\\
\bottomrule
\end{tabular}}
\end{table}

\subsection{Approaches for aerial action recognition}
This section of the report discusses current and state-of-the-art approaches to action recognition trained on the aerial or drone datasets discussed in the previous section.

\vspace{3px}
\noindent\textit{Single-frame classification:} 
A number of approaches took advantage of existing 2D classification networks to perform classification of single frames and subsequently fuse the classification outputs of these frames in a video \cite{ERA,AerialActivityInception,UAVaware}. Similar to other tasks, ResNet and InceptionNet are among the most popular networks in such work as \cite{UAVaware,ERA,AerialActivityInception,DIASR}. More advanced networks such as MobileNet and DenseNet have also been used \cite{DDIR,ERA,HarshalaICCV}. Mou \emph{et al.} experimented multiple networks on the ERA dataset \cite{ERA} and showed that DenseNet achieved the best performance for the aerial action recognition task. Multiple fusion approaches have been employed subsequently such as majority votes \cite{AerialActivityInception} or LSTM-based \cite{UAVaware}.

\vspace{3px}
\noindent\textit{Two-stream CNNs:}
A number of researchers have employed two-stream CNNs for aerial action recognition since they are a natural way to combine appearance and motion \cite{OkutamaAction,MOD20,Drone-Action}. These two-stream CNNs can also be extended to improve the representation capacity. Perera \emph{et al.} employed a subspace representation called KRP-FS to improve representation \cite{MOD20}. Perera \emph{et al.} employed a pose network to detect body keypoints, cropping the input into multiple patches to focus explicitly on body parts for more accurate recognition \cite{Drone-Action,UAV-GESTURE}. Body keypoints can also be used as inputs for a conventional classifiers such as SVM for aerial action classification \cite{AVI}.

\vspace{3px}
\noindent\textit{3D CNNs:}
3D CNNs are still the most popular networks for aerial action recognition. Among the modern networks, I3D \cite{I3D} has been widely adopted for aerial action recognition \cite{UAV-Human,ERA,Game-Action,NEC-Drone,TinyVIRAT}. C3D \cite{C3D} has also been utilized for aerial action recognition \cite{ERA,MEVA}. Others have also upgraded existing 2D networks such as Inception-ResNet \cite{InceptionNet,ResNet} with 3D convolutions to make the suitable for video processing \cite{FullyAuto}. 
%Algamdi \emph{et al.} coupled a 3D CNN with a capsule layer called BVC to improve the modeling capacity \cite{Dronecaps}. 
Mou \emph{et al.} experimented multiple 3D CNNs, \emph{i.e.} C3D , I3D, P3D, TRN, on the ERA dataset \cite{ERA} and showed that TRN \cite{TRN} achieved the best action recognition \cite{ERA}. Ding \emph{et al.} \cite{LightWeight} replaced the backbone network of TSN \cite{TSN} with a lightweight network, \emph{i.e.} MobileNet \cite{MobileNet}, coupled with a focal loss and the modern self-attention mechanism in Transformer \cite{Transformer}.

%\vspace{3px}
%\noindent\textit{Skeleton-based}\\
%\cite{UAV-Human}

\subsection{Techniques to solve aerial action recognition challenges}
The distinct aerial challenges discussed in Section~\ref{sec:aerial_actionrec_challenges} constitute a large number of fine-grained domains, across which an effective aerial action recognition model has to stay robust.

\vspace{-3px}
\subsubsection{Low-resolution}
The most challenging factor in aerial action recognition is the low resolution or small size of objects due to the high flying altitude. A human may appear as tiny as ten pixels in an aerial video. For example, the size of humans in the TinyVIRAT dataset \cite{TinyVIRAT} only accounts for less than 0.01\% of a video spatial size, compared with 0.15\% in the classic ground-based UCF-101 dataset \cite{UCF101}. The most explicit approach for the low resolution challenge is super-resolution \cite{SRsurvey}. Demir \emph{et al.} \cite{TinyVIRAT} proposed to employ a super-resolution network first to increase the resolution of low-resolution aerial videos before action classification. They proposed a dedicated foreground branch in parallel to the video super-resolution network to better focus on foreground objects and humans, which are more critical for accurate action recognition. The proposed approach improved the F1 score of the baseline I3D \cite{I3D} from 28.73\% to 34.49\% on the aerial dataset TinyVIRAT \cite{TinyVIRAT}.

\vspace{-3px}
\subsubsection{Lack of data}
Due to the high cost of capturing and labeling large scale aerial videos with diverse actions, the aerial action recognition community has addressed this challenge by domain adaptation \cite{NEC-Drone} and knowledge distillation \cite{LightWeight}. Choi \emph{et al.} \cite{NEC-Drone} proposed domain adaptation approaches to leverage existing annotated action datasets and unannotated aerial videos. For same source and target label datasets, they proposed unsupervised domain adaption with a cross-entropy loss for classification and a domain adversarial loss for cross-domain learning. For different source and target datasets, the authors employed a triplet loss instead of the cross-entropy loss. Domain adaptation from UCF-101 \cite{UCF101} would increase the accuracy of the I3D baseline up to 70\% on the aerial dataset NEC-Drone \cite{NEC-Drone}. Ding \emph{et al.} \cite{LightWeight} proposed to use I3D \cite{I3D} pre-trained on the Kinetics dataset as a teacher network and learned a student network to fit both the softmax output of the teacher network and the aerial data labels. This knowledge distillation scheme improved the recognition accuracy of the I3D baseline from 78.85\% to 87.81\% on their own UAV action recognition dataset \cite{LightWeight}.

\vspace{-3px}
\subsubsection{Viewing angle}
The unique top and vertical views of aerial footage are challenging for action recognition. To deal with the lack of these views, Sultani \emph{et al.} \cite{Game-Action} proposed to use game engines such as GTA-5 and FIFA to synthesize action videos where multiple views of one action can be easily generated. The authors also proposed using conditional Wasserstein Generative Adversarial Networks (WGANs \cite{WGAN}) to generate additional aerial images from real ground features. Additional images with aerial views help to improve the recognition accuracy of the baseline from 49.7\% to 68.2\%.

\vspace{-3px}
\subsubsection{Fish-eye cameras}
Fisheye cameras are widely used in UAVs due to their ultra-wide view angles; however, their 
%ultra-wide view 
angles would cause severe distortion in videos, which is much more challenging for action recognition than in conventional RGB cameras. Inspired by Spatial Transformer Network \cite{STN}, Li \emph{et al.} \cite{UAV-Human} proposed a guided transformer module, GT-Module, which can be inserted immediately before maxpooling layers in the original I3D \cite{I3D} to warp ``pixel'' in the featuremap extracted from fisheye distorted aerial videos by learning a series of unbounded transformation. This scheme improved the action recognition accuracy on fisheye aerial videos from 20.76\% to 23.24\% on the UAV-Human aerial dataset.

\section{AN OUTLOOK: AERIAL SURVEILLANCE IN NEXT ERA}
\label{sec:GapsOutlook}

%\textcolor{red}{Please help me here}

\subsection{Challenges and Under-investigated Tasks}
\begin{figure}
    \centering
    \includegraphics[width=0.9\columnwidth]{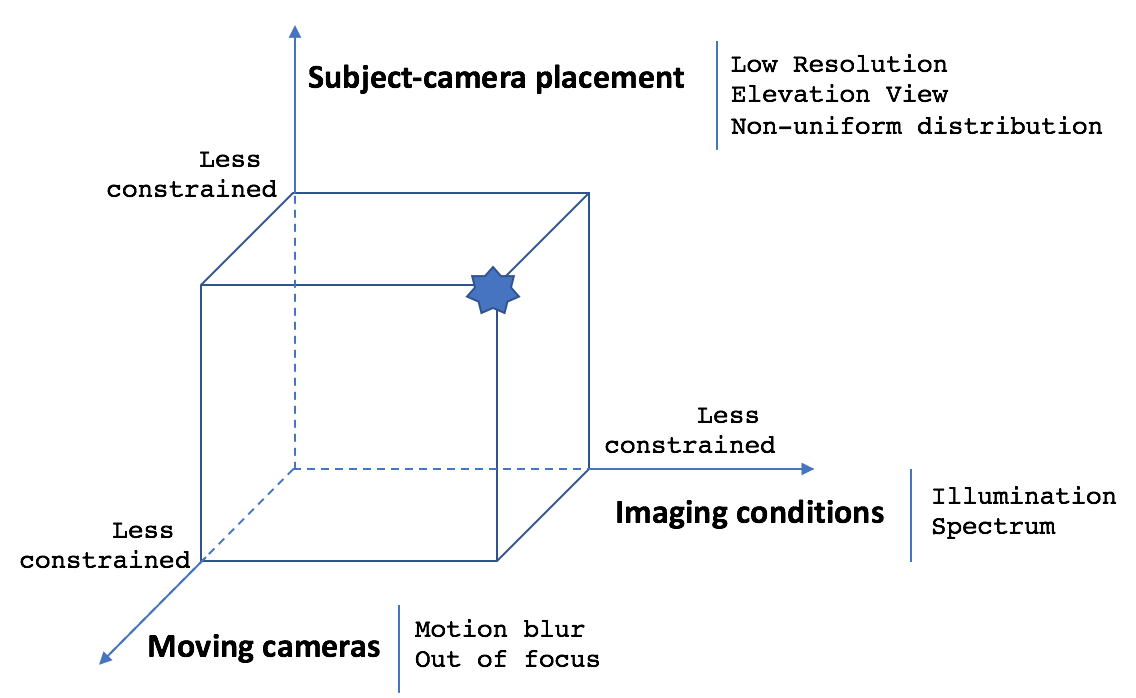}
    \caption{Three challenge dimensions of aerial surveillance: low resolution, vertical perspective, and moving cameras. }
    \label{fig:Aerial_Challenges_Dimensions}
\end{figure}

To fully exploit the potential of aerial surveillance, the challenges of unconstrained and uncooperative environment have to be addressed. The high flying altitudes of airborne platforms on which surveillance cameras and sensors are mounted results in unique settings of aerial surveillance compared to conventional ground-based surveillance in regards to subject-camera placement, imaging conditions, and moving cameras, as illustrated in Fig.~\ref{fig:Aerial_Challenges_Dimensions}. These settings cause the key challenges in small resolution, elevation view, non-uniform distribution, illumination, motion blur, and out of focus data. In the in-depth analysis in four surveillance tasks, we have shown that generic networks drop the performance drastically when shifting to the aerial data. Early approaches have been attempted to tackle these challenges to improve the performance of the aerial surveillance tasks. 

%Fig. compare performance drop of generic networks and the gains from state-of-the-art approaches in all aerial tasks

%Missing tasks: aerial gait recognition, no public dataset and works

Our review of existing work has shown that four aerial surveillance tasks, \emph{i.e.} detection, tracking, re-id and action recognition, have received most attention since they do not require the fine details of human of interest in the scene to perform well. There are a wide range of resources, including public datasets, papers and even competitions available addressing these tasks. In contrast, the human identification task is the least investigated task arising from the small size of face biometric when captured from an aerial platform. The task of recognizing faces from long distance with cameras in a moving aerial platform is extremely challenging and requires further significant research efforts to achieve satisfactory operational performance. On the other hand we make a somewhat surprising observation that the relatively easier task of gait precognition from aerial data  has received significantly less attention.  

\subsection{Open Issues}
We discuss the open issues from two perspectives: data and model development and model deployment.

\subsubsection{Data} 
From a data perspective, three key aspects needing further research are: scale, diversity, and annotation of data.

\vspace{3px}
\noindent\textbf{Large-scale Aerial Surveillance Data:} ImageNet has demonstrated the great importance of leveraging large, comprehensive, and challenging benchmarks to advance computer vision tasks. There is also a critical need for \textit{large-scale public datasets} in the aerial surveillance tasks. More large-scale accurately annotated datasets akin to VisDrones and UAV-Human with more diversified imaging conditions will help to boost the attention and research.

\vspace{3px}
\noindent\textbf{Heterogeneous Aerial Surveillance Data:} Current data collections from aerial platforms have been mainly using monocular RGB data  even though there has been a few multi-modal collection of RGB and thermal. To address the main challenges of aerial surveillance including the reduced number of pixels per sq.cm, extreme view angles, poor visibility, and motion blur, multi-modal fusion techniques could be explored. Multiple heterogeneous modalities beyond visible RGB cameras such as radars, LiDARs, multispectral and hyperspectral sensors would generate rich information for surveillance tasks.

\vspace{3px}
\noindent\textbf{Human Annotation Minimization:} Collecting large scale aerial data has been challenging, but labeling a large amount of data compounds this challenge as the process is very expensive and labor-intensive. Besides unsupervised learning techniques, active learning and learning from virtual data can also help to minimize annotation requirement. However it will require additional techniques, \emph{e.g.} domain adaptation and knowledge distillation, to adapt to the new data.

\subsubsection{Model Development}
From a model perspective, there are three key aspects that need further research: model architecture, effects of unconstrained data on model trustworthiness, prediction uncertainties and model explainability.

\vspace{3px}
\noindent\textbf{Model Architecture:} Many new network architectures such as Transformers \cite{TransformerViT} and Res2Net \cite{Res2Net}, are emerging through computer vision tasks. It would be interesting to explore their performance in the aerial surveillance domain.

\vspace{3px}
\noindent\textbf{Effects of Unconstrained Data on Model Trustworthiness:} The unconstrained nature of aerial surveillance settings results in low quality data with noise. In the current literature, there is no work to report their performance on adverse conditions of blur arising from atmospheric turbulence and motion, out of focus, low light, illumination and weather conditions such as rain and fog. 
%Robustness to these adverse conditions is of paramount importance for real world deployment of aerial surveillance systems. 
The ramifications of these unconstrained data acquisition conditions are further accentuated by the large stand-off distances and the uncooperative nature of the human targets under surveillance. 
%The long range in aerial settings reduces significantly number pixels in the area of interest (\emph{e.g.} the face). These artifacts arising from the unconstrained acquisition of the input data calls for the development of novel techniques to combat these challenges. 
Conventional image and video enhancement techniques would be of limited value in aerial surveillance due to the high levels of data impairments and further focused research efforts in this direction is needed to  mitigate the effects of unconstrained data on model trustworthiness. The recent open call in 2021 by the US government \cite{BRIAR} for the development of BRIAR is an example of the importance of the need for focused research to unlock the full capabilities and potential of aerial surveillance.  

\vspace{3px}
\noindent\textbf{Adversarial Attacks on Aerial Surveillance Tasks:} Recent research has proven that machine learning models are vulnerable to adversarial attacks, both digitally and physically. Understanding the robustness of aerial surveillance models against these attacks, \emph{e.g.} \cite{AerialAdvAttack}, is critical to the security of these models.

\vspace{3px}
\noindent\textbf{Prediction of Uncertainties in Aerial Surveillance:} While many models are emerging, there does not exist any approach to quantify how certain each model is of its predictions of the output. Two sources of uncertainties are from data and model. For example, out of distribution data usually leads to adversarial attacks \cite{UncertaintyDL}. For high stake application such as aerial surveillance, it is important for a model to notify when the output is not reliable.

\vspace{3px}
\noindent\textbf{Explainability of Models in Aerial Surveillance:} Most current surveillance models in general and aerial surveillance models in particular are opaque and only output a scalar prediction value, there is no method providing users semantically understandable explanations for why the model predicts what it predicts. The blackbox-ness could lead to unexpected and detrimental outcomes. Explainable AI (XAI) techniques such as \cite{ExplainableReID} could help to make these models more transparent and understandable to humans.

\subsubsection{Deployment-aware Model Development} 
It is important to design efficient and adaptive models to address scalability issue for practical model deployment.  

\vspace{3px}
\noindent\textbf{Lightweight models:} Deep learning models usually contain tens to hundreds of millions parameters with hundreds of layers, e.g. ResNet-101 has 44.6M parameters and 347 layers \cite{ResNet}. Designing lightweight models is an important step in deploying resource-constrained computing platforms. Lightweight CNNs employ advanced model compression techniques such as pruning, quantization and knowledge distillation \cite{CompressionAcc} to efficiently trade-off between resource and accuracy, minimizing their model size and computations in term of the number of floating point operations (FLOPs), while retaining high accuracies.

\vspace{3px}
\noindent\textbf{Resource-Constrained Model Design:} Most of the current aerial surveillance systems are designed to work either off-line or data is live streamed and processed at the ground stations. There is an urgent need for them to be run on-board. Onboard computing would enable rapid response to surveillance events, reduce the cost and complexity of communication with ground stations, reduce power consumption and improve autonomy of the aerial surveillance systems. However, the limited resources of onboard computing hardware should be taken into account when designing the onboard models. To deal with this, differential constrained design approaches \cite{ConstrainedIrisNet} should be explored.

%\vspace{3px}
%\subsection{Aerial Surveillance Research Promotion}
%Organizing \textit{competitions \& benchmarks} in the major computer vision and machine learning venues will help to further promote this research area.

%------------------------------------------------------------------------
\section{CONCLUDING REMARKS}
\label{sec:Conclusion}
Human-centric aerial surveillance, with its advantages in scale, mobility, deployment and observation, presents new capabilities to enable surveillance in geographically-difficult conditions. The emerging of recent public datasets and initiatives such as VisDrones has further boosted attention and research from the academic community. However, the research in this area is still very daunting. The physical imaging conditions of airborne sensors leads to new challenges in small resolutions, multiple scales, extreme views, motion  blur, atmospheric turbulence, uniformity of distribution, illumination and noise, and calls for new innovations from the computer vision signal processing and machine learning community. This paper presents the very first comprehensive survey with in-depth analysis to understand the state of four key aerial surveillance tasks: detection, tracking, identification (including re-identification) and behavior analysis. While there are still many open issues which require significant more research to progress, the future outlook of aerial surveillance is very promising.  We believe this survey will provide important guidance for future aerial surveillance research.

\section{Acknowledgment} This research was supported by an Australian Research Council (ARC) Discovery grant DP200101942.

{\small
\bibliographystyle{ieee}
\bibliography{ASAS}
}

\clearpage
\appendix

\begin{appendices}

\section{Airborne platforms and imaging sensors for aerial surveillance}
\label{sec:AerialPlatformsSensors}
A wide range of airborne platforms are available with diverse characteristics in flying ranges and altitudes, endurance, speed, manoeuvrability, payload and vulnerabilities as summarized in Table~\ref{tab:airborneplatforms}. These airborne platforms are deployed with a wide range of sensors. In this paper, we only consider techniques for analyzing data acquired by RGB imaging sensors, which capture data with spatial details to be analyzed by computer vision and deep learning. Imaging sensors are categorized by their spectrum as summarized in Table~\ref{tab:aerialsensors}.

\begin{table*}
\centering
\caption{Airborne platforms for aerial surveillance.}
\label{tab:airborneplatforms}
\begin{tabular}{|l|l|l|l|r|l|l|l|} 
\hline
\textbf{}                                                                        & \textbf{Range}                                                                   & \textbf{Endurance}                                                                            & \textbf{Speed}                                                       & \multicolumn{1}{l|}{\begin{tabular}[c]{@{}l@{}}\textbf{Altitude (km)}\end{tabular}} & \textbf{Manoeuvrability}                                                                     & \textbf{Payload}                                                & \textbf{Vulnerabilities}                                                                     \\ 
\hline
Drones                                                                           & Low                                                                              & Low                                                                                           & Slow                                                                 & \begin{tabular}[c]{@{}r@{}}Low\\ 0.5 km\end{tabular}                                           & High                                                                                         & Very low                                                        & Easily detected                                                                              \\ 
\hline
UAVs                                                                             & \begin{tabular}[c]{@{}l@{}}Low to High\\ (1-1,000km)\end{tabular}                & \begin{tabular}[c]{@{}l@{}}Low to Medium\\ (15 minutes\\ to 20 hours)\end{tabular}            & \begin{tabular}[c]{@{}l@{}}Medium\\ 10-300km/h\end{tabular}          & \begin{tabular}[c]{@{}r@{}}Low\\ 5 km\end{tabular}                                             & High                                                                                         & \begin{tabular}[c]{@{}l@{}}Low\\ (1-150kg)\end{tabular}         & \begin{tabular}[c]{@{}l@{}}Weather-\\ dependent,\\ esp. wind\end{tabular}                    \\ 
\hline
\begin{tabular}[c]{@{}l@{}}Balloons\\ (Free or \\ tethered)\end{tabular}         & \begin{tabular}[c]{@{}l@{}}Low\\ (up to 100km\\ per day)\end{tabular}            & \begin{tabular}[c]{@{}l@{}}Medium\\ (10 or more \\ days)\end{tabular}                         & \begin{tabular}[c]{@{}l@{}}Stationary\\ or very \\ slow\end{tabular} & \begin{tabular}[c]{@{}r@{}}Low\\ 5 km\end{tabular}                                             & \begin{tabular}[c]{@{}l@{}}Very low\\ (wind-\\ dependent)\end{tabular}                       & \begin{tabular}[c]{@{}l@{}}Low to Medium\\ (500kg)\end{tabular} & Easily targeted                                                                              \\ 
\hline
\begin{tabular}[c]{@{}l@{}}Rotary aircraft\\ (manned \\ helicopter)\end{tabular} & \begin{tabular}[c]{@{}l@{}}Medium\\ (300km)\end{tabular}                         & \begin{tabular}[c]{@{}l@{}}Low\\ (3 hours)\end{tabular}                                       & \begin{tabular}[c]{@{}l@{}}Medium\\ (up to \\ 350km/h)\end{tabular}  & \begin{tabular}[c]{@{}r@{}}Medium\\ 10 km\end{tabular}                                         & \begin{tabular}[c]{@{}l@{}}Very High\\ (easy turn and\\ stationary \\ capacity)\end{tabular} & \begin{tabular}[c]{@{}l@{}}Medium\\ (10,000kg)\end{tabular}     & Possible fatalities                                                                          \\ 
\hline
\begin{tabular}[c]{@{}l@{}}Fixed-wing \\ aircraft\\ (manned)\end{tabular}        & \begin{tabular}[c]{@{}l@{}}High\\ (10,000km)\end{tabular}                        & \begin{tabular}[c]{@{}l@{}}Low\\ (15 hours)\end{tabular}                                      & High                                                                 & \begin{tabular}[c]{@{}r@{}}High\\ 20 km\end{tabular}                                           & \begin{tabular}[c]{@{}l@{}}High\\ (but cannot \\ flyas slowly)\end{tabular}                  & \begin{tabular}[c]{@{}l@{}}High\\ (250,000kg)\end{tabular}      & Possible fatalities                                                                          \\ 
\hline
Stratollite                                                                      & High                                                                                 & \begin{tabular}[c]{@{}l@{}}High\\ (months)\end{tabular}                                       & Slow                                                                 & \begin{tabular}[c]{@{}r@{}}High\\ 20 km\end{tabular}                                           & Low                                                                                          & \begin{tabular}[c]{@{}l@{}}Low\\ (50kg)\end{tabular}            &                                                                                              \\ 
\hline
Satellite                                                                        & \begin{tabular}[c]{@{}l@{}}Very High\\ (but has fixed\\ trajectory)\end{tabular} & \begin{tabular}[c]{@{}l@{}}Very High\\ (years but \\ revisit time\\ can be days)\end{tabular} & \begin{tabular}[c]{@{}l@{}}Very High\\ (25,000km/h)\end{tabular}     & \begin{tabular}[c]{@{}r@{}}Very High\\ 100-1,000 km\end{tabular}                               & \begin{tabular}[c]{@{}l@{}}Low\\ (only certain \\ types)\end{tabular}                        & \begin{tabular}[c]{@{}l@{}}Medium\\ (5,000kg)\end{tabular}      & \begin{tabular}[c]{@{}l@{}}Limited availability\\ at specific time\\ and place\end{tabular}  \\
\hline
\end{tabular}
\end{table*}

% (1) https://link.springer.com/referenceworkentry/10.1007/978-90-481-9707-1_6

% (2) Table 1 in https://www.mdpi.com/1424-8220/19/3/648/htm

\begin{table*}
\centering
\caption{Imaging sensors for aerial surveillance.}
\label{tab:aerialsensors}
\begin{tabular}{l|l|l|l|l|l|l|l|l|l} 
\hline
                                                                                     \textbf{Sensor Type}                & \textbf{Categories} & \textcolor[rgb]{0.2,0.2,0.2}{\textbf{Wavelength}} & \textbf{Resolution} & \textbf{Distance}                                                                          & \textbf{Visibility} & \begin{tabular}[c]{@{}l@{}}\textbf{Weather}\\\textbf{Affected}\end{tabular} & \textbf{Passive}                           & \textbf{Output} & \textbf{Cost}  \\ 
\hline
\multirow{3}{*}{{Electro-Optical}} & Visible             & \textcolor[rgb]{0.2,0.2,0.2}{0.39-0.75um}         & High                & Medium                                                                                     & Medium              & High                                                                        & \begin{tabular}[c]{@{}l@{}}\\\end{tabular} & 2Dx3            & Low            \\ 
\cline{2-10}
                                                                                                     & Infrared            & 5-14um                                            & Low                 & Long                                                                                       & Medium              & Low                                                                         & \begin{tabular}[c]{@{}l@{}}\\\end{tabular} & 2Dx1            & Medium         \\ 
\cline{2-10}
                                                                                                     & Hyperspectral       & 0.4-2.5um                                         &                     &                                                                                            & Medium              & Medium                                                                      & \begin{tabular}[c]{@{}l@{}}\\\end{tabular} & 2DxM            & Very high      \\ 
\hline
\multirow{2}{*}{{Radio-Wave}}                                    & Radar               & 1mm-300cm                                         & Low                 & \begin{tabular}[c]{@{}l@{}}Very Long\textasciitilde{}\\\textasciitilde{}100km\end{tabular} & Low                 & Low                                                                         & x                                          & 2Dx1*           & Medium         \\ 
\cline{2-10}
                                                                                                     & LiDAR               & 0.9-1.5um                                         & Medium              & \begin{tabular}[c]{@{}l@{}}\\\end{tabular}                                                 & High                & Medium                                                                      & x                                          & 3Dx1            & Very high      \\
\hline
\end{tabular}
\end{table*}

\section{Details on aerial human detection datasets}
\label{sec:aerial_humandetection_datasets}

\vspace{3px}
\noindent\textbf{VisDrone 2020}\\ 
The VisDrone dataset \cite{VisDrones} is collected by the AISKYEYE team at Tianjin University, China. This dataset is behind three ``Vision meets drones'' challenges in ECCV 2018, CVPR 2019 and ECCV 2020. The benchmark dataset consists of 400 video clips formed by 265,228 frames and 10,209 static images, captured by various drone-mounted cameras, covering a wide range of aspects including location (14 different cities separated by thousands of kilometers), environment (urban and country), objects (pedestrian, vehicles, bicycles, \emph{etc.}), and density (sparse and crowded scenes). The dataset was collected using various drone platforms (\emph{i.e.} drones with different models), in different scenarios, and under various weather and lighting conditions. These frames are manually annotated with more than 2.6 million bounding boxes or points of targets of frequent interests, such as pedestrians, cars, bicycles, and tricycles. Some important attributes including scene visibility, object class and occlusion, are also provided for better data utilization.

\vspace{3px}
\noindent\textbf{TinyPersons 2020} \\
The TinyPerson dataset \cite{TinyPersons} is collected from Internet. This dataset is behind two ``Tiny Object Detection'' challenges in ICCV 2019 and ECCV 2020. The unique characteristic of this dataset is the tiny resolution of humans. A majority of human instances appear as tiny as [2,20] pixels, and as small as [20,32] pixels. In total, there are 72,651 objects with bounding boxes have been manually annotated. The dataset is split into a training set, which has 794 labeled images with 42,197 annotations, and a testing set, which has 816 labeled images with 30,454 annotations.

\vspace{3px}
\noindent\textbf{AU-AIR 2020} \\
The AU-AIR dataset \cite{AUAIR} is the first multi-modal UAV dataset for object detection collected by Aarhus University, Denmark. It meets vision and robotics for UAVs having the multi-modal data from different on-board sensors (\emph{i.e.} visual, time, location, altitude, IMU, velocity). The AU-AIR dataset has more than 2 hours raw videos, with 32,823 labeled frames and 132,034 object annotations from 8 object categories related to traffic surveillance. Each frame is also labeled with time, GPS, IMU, altitude, linear velocities of the UAV.

\vspace{3px}
\noindent\textbf{BIRDSAI 2020}\\
The BIRDSAI dataset  \cite{BIRDSAI} is a long-wave thermal infrared dataset containing nighttime images of animals and humans in Southern Africa. The dataset allows for benchmarking of algorithms for automatic detection and tracking of humans and animals with both real and synthetic videos. There are 48 real aerial TIR videos and 124 synthetic aerial TIR videos (generated with AirSim), for a total of 62k and 100k images, respectively. The dataset breaks these into labels of animals or humans, and also provides species information when possible, including for elephants, lions, and giraffes. Information about noise and occlusion for each bounding box is also included.

\vspace{3px}
\noindent\textbf{StanfordDrones 2016}\\
The StanfordDrones dataset \cite{StanfordDrones} is collected by Stanford University, originally aimed to predict human trajectory in crowded scenes but the labels can be employed for aerial human and vehicle detection. The dataset consists of eight unique scenes, recorded in a university campus. The dataset comprises more than 19K targets consisting of 11.2K pedestrians, 6.4K bicyclists, 1.3k cars, 0.3K skateboarders, 0.2K golf carts, and 0.1K buses. 

\vspace{3px}
\noindent\textbf{UAV123 2016}\\
The UAV123 dataset \cite{UAV123} is collected by King Abdullah University of Science and Technology, Saudi Arabia, originally aims for low altitude UAV target tracking. There are 123 videos with 112,578 fully annotated frames. The UAV123 dataset contains 3 subsets: (i) Set1 contains 103 sequences captured using an off-the-shelf professional-grade UAV (DJI S1000) following different objects at altitudes varying between 5–25 m. (ii) Set2 contains 12 sequences captured from a boardcam (with no image stabilization) mounted to a small low-cost UAV following other UAVs. (iii) Set3 contains 8 synthetic sequences captured by the proposed UAV simulator.

\vspace{3px}
\noindent\textbf{Mini-drone 2015}\\
The Mini-drone dataset \cite{MiniDrones} is collected by the Ecole Polytechnique Fédérale de Lausanne (EPFL), originally aimed for drone-based surveillance, helping in managing parking spaces, controlling crowds and reporting useful information such as suspicious behaviors, mis-parked cars, number of free parking spots, \emph{etc.} The created dataset consists of 38 different contents captured in full HD resolution, with a duration of 16 to 24 seconds each, shot with the mini-drone Phantom 2 Vision+ in a parking lot. The dataset contents can be clustered in three categories: normal, suspicious, and illicit behaviors. The low altitude and high resolution of videos are sufficient for fody silhouette, face detection, vehicle license plate and accessories such as bags, backpack, sunglasses, hat, wallet or bottle.

\vspace{3px}
\noindent\textbf{HERIDAL 2018}\\
The HERIDAL database \cite{HERIDAL,HERIDAL2} is collected to support search and rescue missions using drones. It contains over 68,750 image patches of wilderness acquired from an aerial perspective, 29,050 positive samples containing person as well as 39,700 negative samples.Of these, approximately 3,000 image patches are synthetically generated; the others are cropped from real images. Additionally, the HERIDAL database contains approximately 500 labeled, full-size $4,000\times3,000$ pixel real-world images. Full size images could be used for training (Fast or Faster R-CNN) as well as for testing purposes. At the moment 101 images have been selected for testing purposes.

\vspace{3px}
\noindent\textbf{SARD 2021}\\
The SARD database \cite{SARD} is also collected to support search and rescue missions using a DJI Phantom 4A drone. The videos were recorded at a resolution of $1,920\times1,080$ at 50Hz. The drone flew at different altitudes ranging from 5 to 50 meters. The dataset comprises 1,981 manually labeled images from 9 actors.

\vspace{3px}
\noindent\textbf{AgriDrone 2021}\\
AgriDrone is a self-captured data set with focus on person detection in agricultural applications. All 4,586 images have been captured by two different drones: DJI Mavic2 Enterprise and DJI Mavic Pro between Spring and Winter. They share the same resolution of   $3840\times2160$ pixels. The data set is split into 70\% training, 10\% validation and 20\% test data. Due to the rural application area, the average number of humans per image is about two only, which results in a smaller dataset than in the case of the VisDrone data, while the humans are usually bigger.

%-----------------------------------------------------------------------
\section{Details on aerial person tracking datasets}
\label{sec:aerial_tracking_datasets}
\vspace{3px}
\noindent\textbf{UAV123:} UAV123 dataset~\cite{UAV123} contains $123$ fully annotated HD sequences over $110$K frames taken from UAV platforms. Each video has $12$ attribute categories.
%: Aspect Ratio Change (ARC), Background Clutter (BC), Camera Motion (CM), Fast Motion (FM), Full Occlusion (FOC), illumination Variation (IV), Low Resolution (LR), Out-of-View (OV), Partial Occlusion (POC), Similar Object (SOB), Scale Variation (SC), and Viewpoint Change (VC). 
A video may have a variety of attributes by the shooting conditions. The captured targets include \emph{pedestrian}, vehicles, boats, groups and \emph{etc.} The video resolution is between $720$p and $4$K.

% \vspace{3px}
%\noindent\textbf{UAVDT:} UAVDT dataset consists of $10$ hours of raw videos, from which $100$ video sequences of about $80,000$ representative frames are selected.  It has three categories, namely car, truck and bus.

\vspace{3px}
\noindent\textbf{Campus:} The large-scale campus dataset~\cite{StanfordDrones} has images and videos of various classes of targets that move and interact in a real-world university campus. 
The dataset comprises over $19$K targets consisting of $11.2$K \emph{pedestrian}, $6.4$K bicyclists, $1.3$k cars, $0.3$K skateboarders, $0.2$K golf carts, and $0.1$K buses. The resolution of these UAV videos is $1920\times 1080$.

\vspace{3px}
\noindent\textbf{DTB70:} Drone Tracking Benchmark (DTB)~\cite{li2017visual} collects $70$ videos of UAV collected data where the bounding boxes are manually annotated. Some of the videos are captured in a campus, which are for the purpose of tracking \emph{pedestrian}, animals and vehicles. In order to enhance the diversity of the scene, some videos are collected from YouTube. The original resolution of each video frame is $1280\times 720$. 

\vspace{3px}
\noindent\textbf{VisDrone:} The VisDrone team has compiled a dedicated large-scale drone benchmark and organized challenges for tracking, \emph{i.e.} VisDrone-SOT2018~\cite{wen2018visdrone} and VisDrone-SOT2019~\cite{du2019visdrone}. The VisDrone-SOT2018 consists of $132$ videos with $106$K frames. Compared with VisDrone-SOT2018, VisDrone-SOT2019 introduces $35$ new sequences. % ($167$ videos with $189$K frames in total). 
To further increase the diversity of videos and assess the performance of trackers in the wild, VisDrone-SOT2020~\cite{fan2020visdrone} conducts extensive evaluation  of more tracking algorithms using the same dataset in VisDrone-SOT2019. VisDrone2021 further increases the dataset size to $400$ videos with more diverse scenarios.
The annotated targets of VisDrone datasets include \emph{pedestrians}, dogs, bicycles, vehicles, \emph{etc.}  

\vspace{3px}
\noindent\textbf{BIRDSAI:} BIRDSAI~\cite{BIRDSAI} is the first aerial dataset captured by a thermal infrared camera. 
This dataset includes $48$ real videos with a variety of attributes, such as motion blur, large camera motions, background clutter and high altitude. The dataset comprises over $154$K targets consisting of $120$K bounding boxes of wild animals (\emph{i.e.} giraffes, lions, elephants, \emph{etc.}) and about $34$K \emph{human} bounding boxes.

\vspace{3px}
\noindent\textbf{UAVDark135:} UAVDark$135$ dataset~\cite{li2021all} contains fully annotated $135$ videos captured by a standard UAV at night. 
The benchmark includes various tracking scenes, \emph{e.g.} crossings, t-junctions, road, highway, and consists of different kinds of tracked objects like \emph{pedestrian}, boat, bus, car, truck, athletes, house, \emph{etc.} 
To extend the covered scenes, the benchmark also contains some YouTube videos, which are shot on the sea. 
The total frames, mean frames, maximum frames, and minimum frames of the benchmark are $125,466$, $929$, $4571$, and $216$ respectively, making it suitable for large-scale evaluation. 
The videos are captured with the resolution of $1920\times1080$.
at a frame-rate of $30$ frames/s (FPS),

%\vspace{3px}
%\noindent\textbf{UAV-Dataset:}   ~\cite{li2016multi}

%Our dataset comprises of more than100 different top-view scenes for a total of 20,000 targets engaged in various typesof  interactions.  Target  trajectories  along  with  their  target  IDs  are  annotatedwhich makes this an ideal testbed for learning and evaluating models for multi-target  tracking,  activity  understanding  and  trajectory  prediction  at  scale

%-----------------------------------------------------------------------
\section{Details on aerial face recognition datasets}
\label{sec:aerial_facerecognition_datasets}

\vspace{3px}
\noindent\textbf{DroneFace 2017 \cite{DroneFace}:} The DroneFace is an open dataset to simulate the context that a drone seeks lost people on the streets, and tries to recognize the specific target from the air based on the face recognition model established from a few portrait photos. The aerial images are captured by a commercial sports camera (GoPro Hero3+) mounted on a UAV flying at different altitude (1.5, 3, 4, and 5 meters) and captured still images of subjects from 17 meters away from the subjects to 2 meters with 0.5 meters ahead in each step. There are 11 subjects and 2,057 pictures including 620 aerial images, 1,364 frontal face images, and 73 portrait images. The resolution of face regions ranges from $23\times31$ to $384\times384$. Examples of images from this dataset are shown in Fig.~\ref{fig:Aerial_FR_Datasets}.

\vspace{3px}
\noindent\textbf{IJB-S-UAV 2018 \cite{IJB-S}:} The IJB-S dataset is an open-source surveillance video benchmark from the Intelligence Advanced Research Projects Activity (IARPA) to investigate performance of surveillance face recognition. IJB-S stands for IARPA Janus Surveillance Video Benchmark. Aerial facial videos are a part of the dataset. There are 10 UAV videos, captured by a small fixed-wing UAV flying over the collection area, specifically the marketplace, opportunistically capturing surveillance video. There are 5 recognition protocols, one of them is UAV Surveillance-to-Booking, which perform face recognition from UAV video probe to a curated gallery of multiple high resolution mug-shot style photos. Examples of images from this dataset are shown in Fig.~\ref{fig:Aerial_FR_Datasets}.

\vspace{3px}
\noindent\textbf{DroneSURF 2019 \cite{DroneSURF}:} The DroneSURF dataset is a benchmark dataset to investigate performance of face recognition on aerial footage. DroneSURF stands for Drone Surveillance of Faces. The dataset contains 200 videos of 58 subjects, captured across 411K frames, having over 786K face annotations. The proposed dataset demonstrates variations across two surveillance use cases: (i) active and (ii) passive, two locations, and two acquisition times. DroneSURF encapsulates challenges due to the effect of motion, variations in pose, illumination, background, altitude, and resolution, especially due to the large and varying distance between the drone and the subjects. Examples of images from this dataset are shown in Fig.~\ref{fig:Aerial_FR_Datasets}.

%-----------------------------------------------------------------------
\section{Details on aerial person re-ID datasets}
\label{sec:aerial_personreid_datasets}

\noindent\textbf{MRP 2014 \cite{MRP}} Layne \emph{et al.} collected one of the very first aerial datasets for person re-ID. They used a standard remote-operated quadrocopter to capture videos at a resolution of $640\times360$ at 5Hz. Two sets of data were collected. Set 1 contains three flights across an outdoor and indoor environment. These consists of 436, 652, 848 video frames, from which 233, 471, 797 person detections were obtained from 6, 7, 10 distinct people. Set 2 is significantly larger with 6 flights in three unconstrained and heavily crowded outdoor environments. Across each flight, there are between 10k and 30k frames and an average of 8.6k person detections from an unknown number of distinct people. Of this data, they selected 28 uniqre identities and 4,096 images.

\vspace{3px}
\noindent\textbf{DroneHIT 2019 \cite{DroneHIT}} Grigorev \emph{et al.} also employed a standard remote-operated quadrocopter to collect aerial data around a university campus. The drone was flying at a altitude of 25m, capturing videos at a resolution of $1920\times1080$ at 30fps. A total of 101 unique identities are extracted, where each person has about 459 images.

\vspace{3px}
\noindent\textbf{P-DESTRE 2020 \cite{PDESTRE}} Kumar \emph{et al.} collected a comprehensive dataset for re-identification across multiple days with the change in clothing of persons. The authors also provided 16 human attributes of person instances for the person search task. The annotations for the attributes include demographic information: gender, ethnicity and age, appearance information: height, body volume, hair color, hairstyle, beard, mustache; accessories information: glasses, head accessories, body accessories; clothing information and action information. The dataset was collected in two universities using DJI Phantom 4 drones at two altitudes: 5.5 and 6.7 meters. The videos are captured at a resolution of $3,820\times2,160$ at 30fps. There are a total of 269 unique subjects with 14.8M person detections. The re-ID task can be performed based on visual input, \emph{i.e.} image or video, or heterogeneously textual input of attributes. 

\vspace{3px}
\noindent\textbf{PRAI-1581 2020 \cite{PRAI1581}} Zhang \emph{et al.} recently collected a large dataset for aerial person re-ID. The images were shot by two DJI drones at an altitude ranging from 20 to 60 meters. The dataset consists of 39k images of 1581 unique subjects. The resolution of persons is low, ranging from 30 to 150 pixels. The high flying altitude makes the diversity of views, poses more extreme. 

\vspace{3px}
\noindent\textbf{UAV-Human 2021 \cite{UAV-Human}} Li \emph{et al.} just published a new dataset for aerial person re-identification in CVPR 2021. The dataset was collected by a flying UAV in multiple urban and rural districts in both daytime and nighttime over three months, hence covering extensive diversities w.r.t. subjects, backgrounds, illuminations, weathers, occlusions, camera motions, and flying altitudes. The dataset contains videos and annotations for multiple tasks, including action recognition, pose estimation and person re-identification. There are 41,290 frames and 1,144 identities for person re-identification and 22,263 frames for attribute recognition. The unique characteristic of the dataset is multimodal aerial data was captured and provided using a depth sensor (Azure DK), a fisheye camera, and a night-vision camera.

\vspace{3px}
\noindent\textbf{MEVA 2021 \cite{MEVA}} MEVA is a new and very-large-scale dataset for human activity recognition. The collection observed approximately 100 actors performing scripted scenarios and spontaneous background activity over a three-week period at an access-controlled venue, collecting in multiple modalities with overlapping and non-overlapping indoor and outdoor viewpoints. The resulting data includes video from 38 RGB and thermal IR cameras, 42 hours of UAV footage, as well as GPS locations for the actors.

\section{Details on aerial human behavior datasets}
\label{sec:aerial_humanaction_datasets}

% Action datasets
%-------------------------------------------------------------------------------------------------------
\vspace{3px}
\noindent\textbf{UAV-Human 2021 \cite{UAV-Human}} Li \emph{et al.} recently published a new dataset for aerial human behavior understanding in CVPR 2021. The dataset was collected by a flying UAV in multiple urban and rural districts in both daytime and nighttime over three months, hence covering extensive diversities w.r.t. subjects, backgrounds, illuminations, weathers, occlusions, camera motions, and flying altitudes. The dataset contains videos and annotations for multiple tasks, including action recognition, pose estimation and person re-identification. There are 67,428 multi-modal video sequences and 119 subjects for action recognition. The unique characteristic of the dataset is multimodal aerial data was captured and provided using a depth sensor (Azure DK), a fisheye camera, and a night-vision camera.

\vspace{3px}
\noindent\textbf{Drone-Action 2019} \\
The Drone-Action dataset \cite{Drone-Action} was collated by a team of academics out of the University of South Australia, this dataset was developed in order to fill the gap of limited outdoor footage. Most datasets were filled with footage captured in doors. Published in 2019, Drone-Action consists of 10 actors performing 13 different actions; Walking, walking side, jogging, jogging side, running, running side, hitting with bottle, hitting with stick, stabbing, punching, kicking, clapping and waving hands. In total 240 videos were collated, consisting of 66,191 frames annotated with body joint estimations and bounding boxes.

\vspace{3px}
\noindent\textbf{Okutama-Action 2019} \\
The Okutama-action dataset \cite{OkutamaAction} is a comprehensive dataset published in 2019 out of the University of Tokyo. The dataset consists of fully annotated sequences similar to Drone-action this dataset consists of 12 action classes; handshaking, hugging, reading, drinking, pushing/pulling, carrying, calling, running, walking, lying down, sitting and standing. Which are all categorized as either human-human, human-object, or none-interaction. The dataset consists of 43 videos and 77365 frames fully annotated with bounding boxes and body joint estimations. Fig.~\ref{fig:okutama_sample} below depicts a sample from the Okutama dataset.

\vspace{3px}
\noindent\textbf{UCF-ARG:} action features videos collected at an altitude of 400-450ft (122-137m). Collated in 2008, the videos were captured using a camcorder attached to a blimp\cite{UCF_2008}. UCF Aerial action is part of the UCF-ARG dataset which consists of footage collected from cameras stationed in different positions; Aerial, Rooftop and ground. It consists of 48 aerial videos of 12 actors performing 10 actions; boxing, carrying, clapping, digging, jogging, open/closing trunk, running, throwing, walking and waving \cite{UCF_2011}.

\vspace{3px}
\noindent\textbf{Game-Action 2020:} A unique approach to collecting aerial videos is the use of video game characters instead of real people. UCF along with a team out of ITU Pakistan introduced the first of its kind dataset, games action dataset, which uses human action footage from video games \cite{Game-Action}. This footage is of characters in Grand Theft Auto 5 (GTA5) performing 7 different human actions; cycling, fighting, soccer kicking, running, walking, shooting and skydiving. The team used the game FIFA to collect the footage of soccer kicking. The dataset consists of 200 videos, 100 of which are aerial footage. Sample of the aerial footage in the dataset is depicted in Fig.~\ref{fig:games_sample} below.

\vspace{3px}
\noindent \textbf{DIASR 2020:} Search and rescue implementations of drone-based technology has garnered little interest from researchers so far, however, due to the increasing capabilities of drones and requirements to monitor remote areas drone surveillance has become popular. To fill the gap of a lack of search and rescue implementations \cite{DIASR} have proposed an search and rescue dataset; Drone image action dataset for search and rescue (DIASR) which consists of 30,000 frames of HD video and images of multiple actors performing 6 different actions.The actions performed are intended to mimic a person signaling / asking for help and include; waving, standing, sitting, laying and handshaking. As the name suggests the primary reason this dataset was curated was to cultivate the implementation of search and rescue implementations of drones using computer vision. A sample of the dataset is depicted in ~Fig. \ref{fig:SAR_dataset} below.

\vspace{3px}
\noindent \textbf{AVI 2020:} The AVI dataset \cite{AVI} was developed to identify violent individuals in public areas. The complete datasets consist of 2,000 images with 10,863 humans where 5,124 (48\%) engaged in one or more of the five violent activities of (1) Punching, (2) Stabbing, (3) Shooting, (4) Kicking, and (5) Strangling. These activities are performed by 25 subjects between the ages of 18–25 years. These images are recorded from the parrot drone at four heights of 2m, 4m, 6m and 8m.

\vspace{3px}
\noindent \textbf{Youtube Aerial 2020:}

% Interaction datasets
%-------------------------------------------------------------------------------------------------------

\vspace{3px}
\noindent\textbf{UT-interaction 2009}\\
Published in 2009, UT-interaction \cite{UT-Interaction} is a dataset containing 6 different actions similar to Drone-Action; shake-hands, point, hug, push, kick and punch. This dataset is not as comprehensive as Drone-Action, UT-interaction is comprised of 20 videos, shot at 720x480 resolution.

\begin{figure}
\centering
\includegraphics[width=\columnwidth]{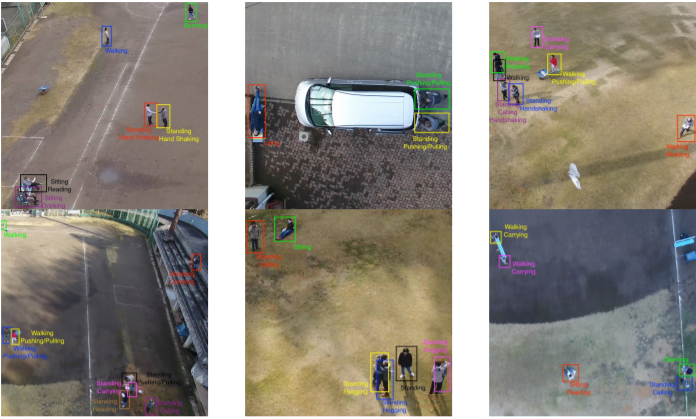}
\caption{Sample frames from the Okutama-action dataset \cite{OkutamaAction}.}
\label{fig:okutama_sample}
\end{figure}

\begin{figure}
\centering
\includegraphics[width=\columnwidth]{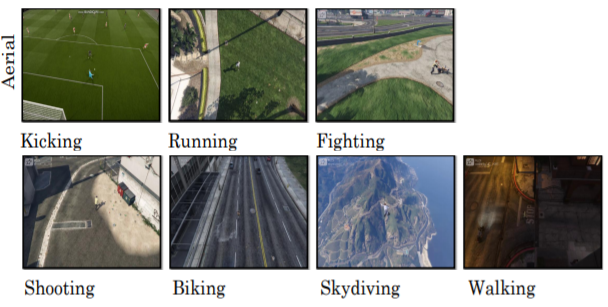}
\caption{Sample from the games-action dataset \cite{Game-Action}.}
\label{fig:games_sample}
\end{figure}

\vspace{3px}
\noindent\textbf{MOD 2020:} is another dataset produced by the team at the University of South Australia, it is their most recent dataset. The dataset is curated to combat the challenges of the angle / viewpoint of aerial footage and for this reason the dataset fills a gap in the current literature regarding a scarcity of multi viewpoint datasets. The dataset follows in the footsteps of the others in that it is specifically focused on outdoor scenes of action recognition. The dataset is a collection of 2324 video clips, 503,086 frames, sourced from YouTube or recorded on an unspecified drone by the team. \cite{MOD20}.

% Gesture datasets
%-------------------------------------------------------------------------------------------------------

\vspace{3px}
\noindent\textbf{UAV-Gesture 2018} \\
The same team out of the University of South Australia produced another dataset, UAV-Gesture \cite{UAV-GESTURE}. The aim of this dataset was that action recognition datasets where aimed specifically at actions similar to those discussed in the previous dataset. UAV-Gesture was designed to fill the gaps in command signaling/gesturing datasets, the team found there was a lack of data collected in an outdoor environment. Published the year prior to Drone-Action in 2018, UAV-Gesture comprises of 10 actors performing 13 different gestures, all of which are aimed at being able to control/command the drone to maneuver in a programmed way upon recognition of the action. The dataset contains 119 videos with 37151 frames annotated with bounding boxes and body joints estimations.

\vspace{3px}
\noindent \textbf{DDIR 2020:} Systems that implement human action recognition are typically focused on identifying multiple different actions, such as all the datasets highlighted in this section so far. In contrast to this \cite{DDIR} have proposed an approach that builds a natural interaction system to guide autonomous drones. The approach focuses on fine grained variations of the same human action; pointing and in order to validate their framework, and foster future innovation in the same area they have curated a dataset with variability in user appearance, viewpoint camera distance and scenery. The dataset presented is called Direction Dataset for Interaction with Robots (DDIR) and is divided into 5 subsets denoted DDIR1 to DDIR5. The number of actors range from 5 to 7 with up to 22,628 frames of VGA and FWVGA video. Recorded on an unspecified drone, the dataset consists of 26 different actions, representing a different direction.

% Group activity datasets
%-------------------------------------------------------------------------------------------------------

%\vspace{3px}
%\noindent\textbf{ERA 2018} 

\begin{figure}
\centering
\includegraphics[width=\columnwidth]{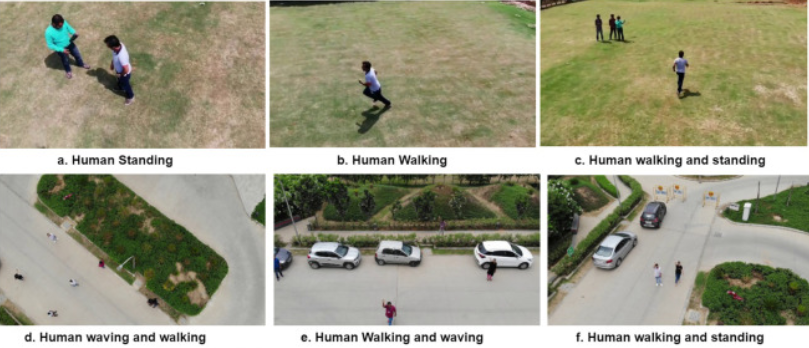}
\caption{Sample from the DIASR dataset \cite{DIASR}.}
\label{fig:SAR_dataset}
\end{figure}

\vspace{3px}
\noindent \textbf{DLR’s Aerial Crowd Dataset (DLR-ACD):} 
The DLR-ACD dataset \cite{MRCNet} is a collection of aerial images for crowd counting and density estimation, as well as for person localization at mass events. It contains 3 large RGB aerial images withaverage size of $3619\times5226$ pixels acquired through 16 different flight campaigns at various mass events and over urban scenes involving crowds, such as sport events, city centers, open-air fairs and festivals.

The images were recorded using a camera system composed of three standard DSLR cameras (a nadir-looking and two side-looking cameras) mounted on an airborne platform installed on a helicopter flying at an altitude between 500 m to 1600 m with spatial resolution (or ground sampling distance – GSD) ranges from 4.5 to 15 cm/pixel. The dataset was labeled manually with point-annotations on individual people and contains 226,291 person annotations in total, ranging from 285 to 24,368 annotations per image. 

\end{appendices}

%\section{Network architecture}

%FeatNet, DRFNet, ResNet, ComplexIrisNet

\end{document}